\documentclass[twoside]{article}

\usepackage[preprint]{aistats2027}

\usepackage[round]{natbib}

\usepackage{bm}
\usepackage{amssymb}
\usepackage{hyperref}
\hypersetup{
    colorlinks=true,
    citecolor=green,
    linkcolor=blue,
    urlcolor=blue
}

\usepackage{graphicx}
\newtheorem{proposition}{Proposition}
\usepackage{makecell}

\begin{document}



% \twocolumn[

% \aistatstitle{Anomaly Detection via Density Estimation of Autoencoder Composite Features}

% \aistatsauthor{ Yalin Liao \And ustin J. Brockmeier}

% \aistatsaddress{ Institution 1 \And  Institution 2} ]

\onecolumn

\title{Anomaly Detection via Density Estimation of Autoencoder Composite Features}

\author{
Yalin Liao \qquad Austin J. Brockmeier
\\[0.5em]
\parbox{\textwidth}{\centering
Department of Electrical and Computer Engineering \\
University of Delaware \\
Newark, Delaware \\[0.3em]
\texttt{yalin@udel.edu} \qquad \texttt{ajbrock@udel.edu}
}
}

\date{}

\maketitle

\date{}

\maketitle

\date{}

\maketitle

\begin{abstract}
Autoencoders (AEs) are widely used for anomaly detection (AD), either by directly using reconstruction errors as anomaly scores or by learning compact latent representations for subsequent AD, such as density estimation in the latent space. We demonstrate that latent features extracted by AEs alone can be insufficient for anomaly detection. We therefore propose to complement the latent features with reconstruction errors to construct a composite feature representation and employ noise contrastive estimation (NCE) to estimate its density for anomaly detection. We further establish a connection between joint density modeling of the composite features and the $\beta$-VAE objective. To encourage the estimated density to decrease with increasing reconstruction errors, we propose a novel augmentation strategy that generates artificially lower reconstruction errors. Experiments across multiple benchmark datasets demonstrate that the proposed approach achieves performance comparable to state-of-the-art AD methods.
\end{abstract}

\section{Introduction}
The goal of anomaly detection (AD) is to identify observations that deviate from a given typical distribution \citep{chandola2009anomaly}. Along with the overall rise of deep learning, neural network-based anomaly detectors are often used for image distributions \citep{seebock2016identifying, ruff2018deep, sabokrou2018deep}. Autoencoders (AEs), often with convolutional architectures, are trained on the `normal' data and widely applied for AD in one of two distinct cases. In the first case, the reconstruction error of an instance serves as the anomaly score \citep{sakurada2014anomaly, xia2015learning, chen2018autoencoder}. In the second case, the AE's learned latent representation of the data in the bottleneck layer are treated as features, and subsequently, a machine learning or statistical approach is employed to detect anomalies based on this learned representation \citep{andrews2016detecting,  sabokrou2018deep, xu2015learning}.

\begin{figure*}[t]
\centering
\includegraphics[width=0.95\textwidth]{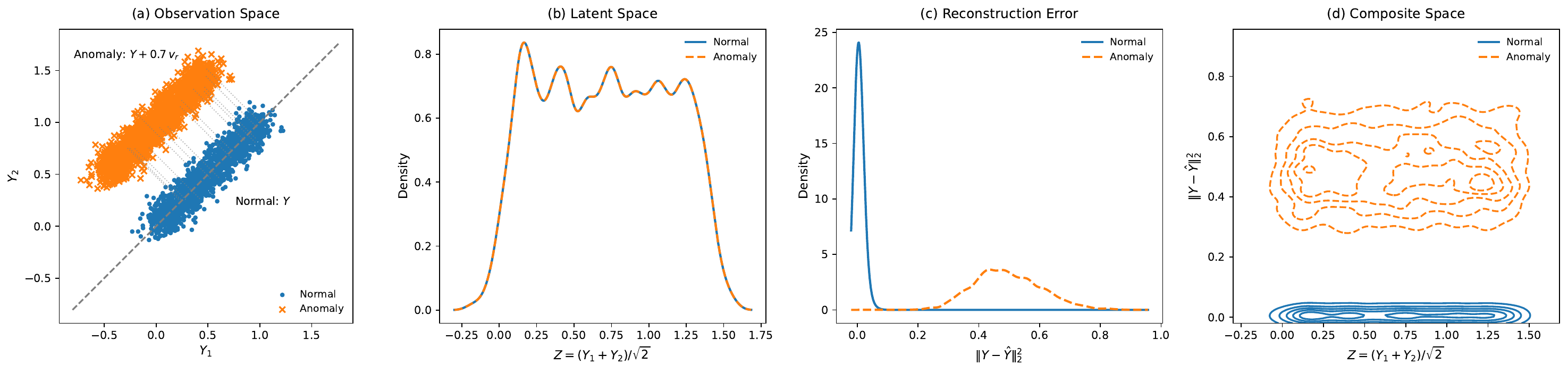}
\caption{Illustration of Proposition 1 with normal observations $Y=(Y_1, Y_2)^T$, where $Y_i=X+\epsilon_i, X\sim U([0,1])$, and $\epsilon_i\sim\mathcal{N}(0, 0.08^2)$. Anomalies are generated by perturbing $Y$ by $0.7v_r$ along the residual direction $v_r=(-1,1)^T/\sqrt{2}$. The panels show the observations, latent representations, and reconstruction errors.}
\label{Illustrative_example}
\end{figure*}

However, latent representations alone may be insufficient for AD. Normal and anomalous instances can overlap in the latent space but differ substantially in reconstruction errors. Figure~\ref{Illustrative_example} provides a simple illustration of this
phenomenon. The theoretical basis for this example is established in the
following proposition, which characterizes the optimal one-dimensional
representation and its corresponding reconstruction. The proof is provided
in Appendix \ref{proof-motivation}.
%%%%%%%%%%%%%%%%%%%%%%%%%%%%%%
%%%%%%%%%%%%%%%%%%%%%%%%%%%%%
\begin{proposition}
\label{prop-motivation} Let $X$ be a random variable with $0<\operatorname{Var}(X)<\infty$, and let $Y=(Y_1,Y_2)^T$, where $Y_i=X+\epsilon_i, \epsilon_i\sim\mathcal{N}(0,\sigma^2)$, and $X,\epsilon_1,\epsilon_2$ are mutually independent. For a one-dimensional encoder $Z=h(a^TY+b)$ with injective $h$ and an arbitrary nonlinear decoder $D$, the minimum reconstruction error
$
\min_{a,b,h,D}
\mathbb{E}\!\left[
\left\|Y-D\!\left(h(a^TY+b)\right)\right\|^2
\right]
$
is attained by the latent representation
$Z^\ast=h\left(\frac{Y_1+Y_2}{\sqrt{2}}\right)$,
up to an invertible reparameterization of $Z^\ast$, with optimal reconstruction
$\hat{Y}^\ast=\left(
\frac{Y_1+Y_2}{2},
\frac{Y_1+Y_2}{2}
\right)^T
$.
\end{proposition}

Motivated by this observation, we jointly model latent representations and reconstruction information through a composite feature. Specifically, given either raw data or features extracted by a pretrained neural network, we learn a low-dimensional representation and reconstruction using an AE. We then augment the latent representation with normalized squared reconstruction error and cosine dissimilarity to construct the composite feature. On this composite feature, we perform density estimation using noise contrastive estimation (NCE)~\citep{gutmann2010noise,gutmann2012noise} and use the negative estimated log-density as the anomaly score.

To further understand this joint modeling, we revisit Latent Space Autoregression (LSA)~\citep{abati2019latent}, an AD method, through the lens of the $\beta$-VAE objective~\citep{higgins2017beta}. In particular, we show that the LSA objective can be reformulated as a  $\beta$-VAE objective for modeling the joint distribution of the latent features and reconstruction error under two conditions: the prior distribution of the latent features is replaced by an autoregressive model, and the conditional distribution of the reconstruction error given the latent features is specified by an exponential-type likelihood. This reformulation shows that LSA can be interpreted as modeling the joint distribution of latent features and reconstruction error under specific distributional assumptions. Our approach naturally generalizes this perspective by estimating the joint density of the composite feature. We employ NCE for density estimation owing to its relatively simple training procedure, minimal distributional assumptions, and limited hyperparameter requirements.

For effective density estimation, we follow the established practice of optimizing the noise distribution to better match the data distribution. To further adapt NCE to the AD setting, we augment augment the training set by introducing additional normal instances with artificially constructed reconstruction features. Specifically, these instances are constructed to ensure that the marginal density for low reconstruction errors is no less than the corresponding noise density. This reduces the risk of normal samples being misclassified as anomalous. The resulting framework performs NCE-based density estimation on the composite feature using both the original and augmented normal samples and is referred to as Composite Feature Augmented NCE (\textbf{CANCE}).

Experiments across multiple benchmark datasets show that CANCE achieves competitive performance with state-of-the-art AD methods. Ablation studies validate the effectiveness of the composite-feature formulation and the proposed sample augmentation strategy.

\section{Related Work}
Density-based AD identifies anomalies as samples with low probability under the estimated distribution of normal data \citep{abati2019latent}. Classical approaches include Kernel Density Estimation (KDE) \citep{parzen1962estimation} and Gaussian mixture models (GMMs) \citep{bishop1994novelty}, but their effectiveness deteriorates in high-dimensional spaces due to the curse of dimensionality \citep{ruff2021unifying}. To address this issue, prior work performs anomaly detection on lower-dimensional representations learned by AEs \citep{andrews2016detecting,  sabokrou2018deep, xu2015learning}. Alternatively, reconstruction-based methods identify anomalies through their larger AE reconstruction errors \citep{sakurada2014anomaly, xia2015learning, chen2018autoencoder}. DAGMM \citep{zong2018deep} integrates both latent and reconstruction features into a composite representation and jointly learns the representation and its density model using an AE and a GMM. In contrast, we decouple representation learning from density estimation and apply NCE to the composite features, requiring fewer distributional assumptions and hyperparameters.

To model more complex distributions of normal data, deep generative models have been increasingly explored for AD. Energy-based models \citep{zhai2016deep} rely on computationally expensive MCMC sampling, while autoregressive \citep{salimans2017pixelcnn++} and flow-based models \citep{kingma2018glow} enable direct likelihood estimation but may assign high likelihoods to anomalies \citep{choi2018waic,ren2019likelihood,yoon2024energy}. GAN-based methods, such as AnoGAN \citep{schlegl2017unsupervised} and EGBAD \citep{zenati2018efficient}, model normal data implicitly, whereas VAEs \citep{kingma2013auto} approximate data likelihood through the evidence lower bound (ELBO), with reconstruction-based scores also explored for AD \citep{an2015variational,xu2018unsupervised}. More recently, diffusion-based methods have shown competitive performance for AD; DTE \citep{livernoche2024diffusion} estimates diffusion time as the anomaly score and substantially improves inference efficiency over DDPM-based AD.

Beyond modeling the distribution of normal data, one-class classification methods learn decision boundaries from normal data. SVDD \citep{tax2004support} encloses normal samples within a hypersphere, while DSVDD \citep{ruff2018deep} extends this approach with deep representations. DROCC \citep{goyal2020drocc} instead learns a boundary around the normal-data manifold. To address potentially unreliable decision boundaries, DPAD \citep{fu2024dense} learns dense low-dimensional representations for density-based AD.

\section{Proposed Methodology}
\begin{figure*}[t]
\centering
\includegraphics[width=0.9\textwidth]{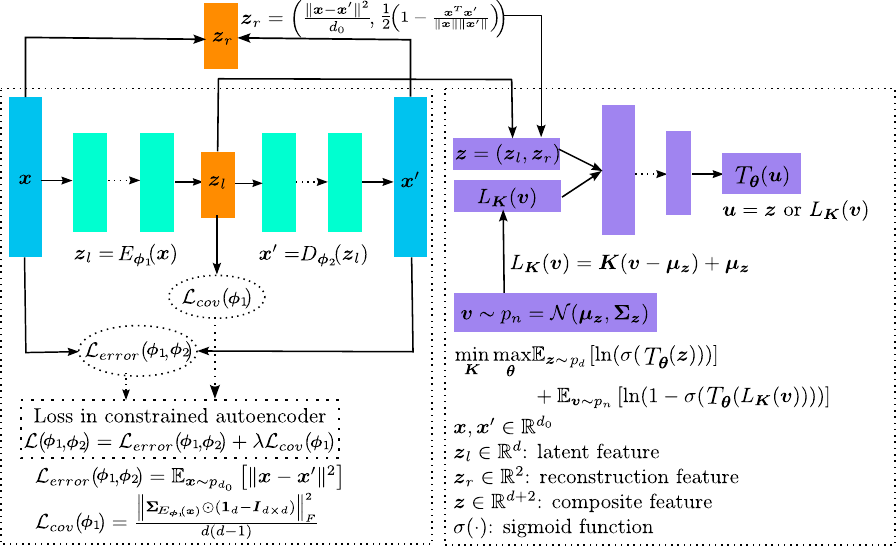}
\caption{CANCE for AD.}
\label{overall method}
\end{figure*}

We present CANCE, a two-step methodology for AD, as illustrated in Fig \ref{overall method}. First, we train an AE to construct a composite feature using specific techniques to ensure the AE latent is appropriate for the subsequent NCE. Subsequently, we utilize a network trained with NCE to estimate the negative log-likelihood function based on the composite feature, which serves as the scoring function, with higher values indicating anomalies and lower values signifying normality.

\subsection{Overview of the Anomaly Score}
For input data $\bm x\sim p_{d_0}$ in the data space $\mathbb{R}^{d_0}$ distributed according to $p_{d_0}$, we propose to estimate an anomaly score function $S:\mathbb{R}^{d_0}\rightarrow \mathbb{R}$ to predict anomalies. Ideally, $S(\bm x)$ could approximate the negative log-likelihood function, but since probability density functions often do not exist in $\mathbb{R}^{d_0}$, especially for image datasets, we estimate a score function $S_C: \mathbb{R}^{d+2}\rightarrow \mathbb{R}$ as the negative log-likelihood of the composite feature $\bm z=\left (E_{\bm\phi_1}(\bm{x}), z_e(\bm{x}), z_c(\bm{x}) \right ) = C(\bm x)\in \mathbb{R}^{d+2}$, consisting of $d$ latent features $\bm z_l \in\mathbb{R}^d$ and two reconstruction quality features $\bm z_r\in\mathbb{R}^2$ (squared error and cosine dissimilarity), where $\bm z_l = E_{\phi_1}(\bm x)$ and $x' = D_{\phi_2}(\bm z_l)$ denote the latent representation and reconstruction, respectively, obtained from an AE applied either directly to the raw input or to features extracted by a pretrained network. The reconstruction features are defined as the normalized squared reconstruction error $z_e(\bm{x})=\frac{\|\bm{x}-\bm{x}'\|^2}{d_o}$ and the cosine dissimilarity $z_c(\bm{x})=\frac{1}{2}\left(1-\frac{\bm x^T\bm x'}{\|\bm x\|\|\bm x'\|}\right)$. The distribution of composite features is denoted as $p_d$. (To simplify notation, distributions will be denoted by their probability density or mass functions for discrete random variables.)

The anomaly score function $S_C$ is derived through noise contrastive estimation (NCE) \citep{gutmann2010noise,gutmann2012noise}. In this process, an estimation network $T_{\bm\theta}(\bm u)$ is trained with supervision to predict whether the network's input $\bm u$ is from the composite feature distribution $p_d$, such that $\bm u =\bm z = C(\bm x) \sim p_d$, or from a specified noise distribution $p_n$,  such that $\bm u = \bm v\sim p_n$. After training, the optimized estimation network $T_{\bm\theta^*}(\bm z)$ approximates the log density ratio $\ln\frac{p_d(\bm z)}{p_n(\bm z)}$ plus a constant, which provides an approximation of the negative log-likelihood $-\ln {p_d(\bm z)}$, since $p_n$ is known. $S_C$ is this approximation, such that a high $S(\bm x)=S_C(C(\bm x))$ suggests the data point $\bm x$ is likely to be abnormal, while a low value indicates normality.

\subsection{Autoencoder Network Design \& Training}
AEs are designed to provide a compressed space of latent features, which can be used for AD as shown in previous work \citep{sakurada2014anomaly, sabokrou2018deep}, and accurate reconstructions of normal data, while ideally providing poor reconstructions of anomalies. However, with vanilla training the latent feature distribution could create a degenerate distribution, i.e., the latent representations could contract to lie in a strict subspace of the whole latent space \citep{ozsoy2022self}, creating a singular covariance matrix. Non-degeneracy of the learned representations is necessary for the subsequent NCE as the contrastive noise is assumed to follow a multivariate Gaussian, and a degenerate distribution that lies in a subspace makes the problem ill-posed.  Thus, the architecture of the AE should be designed such that first and second moments of the latent representation better match a standard Gaussian. Specifically, we implement batch normalization directly before the latent space and penalize correlation among latent features in terms of the squared values of off-diagonal elements in the covariance matrix. This covariance loss encourages a diagonal covariance, and combined with normalization layer, mitigates a singular covariance matrix. This will objectively encourage the development of statistically uncorrelated latent feature with isotropic variance. These enhancements for a deterministic AE echo the goal of disentanglement and Gaussian prior in a $\beta$-VAE~\citep{higgins2017beta}.

Based on these considerations, we train the AE by minimizing
$$\mathcal{L}(\bm\phi_1,\bm\phi_2)=\mathcal{L}_{error}(\bm\phi_1,\bm\phi_2)+\lambda\mathcal{L}_{cov}(\bm\phi_1),$$
 with respect to the encoder and decoder parameters $\bm\phi_1$ and $\bm\phi_2$, respectively, where $\lambda$ controls the trade-off between the reconstruction and covariance losses. The two loss terms are defined as
\begin{align}\label{ae_loss}
\mathcal{L}_{error}(\bm\phi_1,\bm\phi_2)&=\mathbb{E}_{\bm{x}\sim p_{d_0}}
\left[\|\bm{x}-D_{\bm\phi_2}(E_{\bm\phi_1}(\bm{x}))\|^2\right],\\
\mathcal{L}_{cov}(\bm\phi_1)&=\frac{1}{d(d-1)}\left\Vert
\mathrm{off}(\bm\Sigma_{E_{\bm\phi_1}(\bm{x})})\right\Vert_F^2.
\end{align}
Here, $\mathcal{L}_{error}(\bm\phi_1,\bm\phi_2)$ is the mean squared error of the reconstruction, $\mathcal{L}_{cov}(\bm\phi_1)$ is the mean of the squared off-diagonal elements in the covariance matrix $\bm\Sigma_{E_{\bm\phi_1}(\bm{x})}$ of the latent representation $\bm{z}_l=E_{\bm\phi_1}(\bm{x})$, $\mathrm{off}(\bm\Sigma)=\bm\Sigma-\bm\Sigma  \odot \bm{I}_{d}$, $\odot$ is the element-wise product,  and $\bm{I}_{d}$ is the identity matrix.

Additionally, we adopt a decoupled training strategy to mitigate the impact of the additional covariance loss on the AE's ability to reconstruct input images. Specifically, the encoder is updated only during the first stage of training and is subsequently frozen in the second stage while the decoder is further trained. This decoupled training method also facilitates the learning of a higher-quality latent feature \citep{hu2024complexity,loaiza2024deep}.  In summary, we propose to learn structured representations by training a constrained AE to jointly minimize the reconstruction error and a covariance loss term that encourages the components of the latent feature to be statistically uncorrelated.

\subsubsection{Connection to LSA}\label{link2lsa}
Here, we revisit the LSA objective through the VAE framework to establish its connection to density estimation on composite features. The VAE, with latent encoding  distribution $q_{\phi_1}(z_l |\bm x)$ and reconstruction distribution $q_{\phi_2}(\hat{\bm x} |\bm z_l )$ is trained by maximizing the Evidence Lower Bound (ELBO), which is a lower bound on the data log-likelihood $\ln p_{d_0}(\bm x)$ using a prior $p(\bm z)=\mathcal{N}(\bm 0, \bm I)$,  expressed as
\begin{align}\label{elbo}
\mathbb{E}_{q_{\phi_1}\!(\bm z_l |\bm x)}[\ln q_{\phi_2}(\bm x|\bm z_l)] - D_\text{KL}(q_{\phi_1}(\cdot |\bm x)\|p(\cdot ) ) ,
\end{align}
where the first term represents the reconstruction loss, encouraging the model to accurately reconstruct the input data, and the second term encourages the approximate posterior distribution to be close to the prior $p$. Once the VAE is trained to maximize the objective $\mathbb{E}_{p_d(\bm x)} \left [\mathit{ELB}(\bm x)\right ]$, several quantities derived from the model can serve as anomaly scores. These include the model likelihood $\mathbb{E}_{p(\bm z_l)}[q_{\phi_2}(\bm x|\bm z_l)]$, the reconstruction probability,  $\mathbb{E}_{q_{\phi_1}(\bm z_l|\bm x)}[\ln q_{\phi_2}(\bm x|\bm z_l)]$ \citep{an2015variational}, or ELBO itself \citep{abati2019latent}. Typically, the decoder $D_{\phi_2}$ provides the mean of a Gaussian distribution $
q_{\phi_2}(\bm{x}|\bm{z}_l) = \mathcal{N}\left(D_{\phi_2}(\bm{z}_l), \sigma^2 I\right)$. Substituting $\ln q_{\phi_2}(\bm{x}|\bm{z}_l) = c_0 -\frac{1}{2\sigma^2} \|\bm{x} - D_{\phi_2}(\bm{z}_l)\|^2$, with $\sigma = 1/\sqrt{2}$, into  \eqref{elbo}, negating, and ignoring $c_0$ yields
\begin{align}\label{vae_obj}
\mathbb{E}_{q_{\phi_1}\!(\bm z_l|\bm x)}\left[\|\bm{x}\! -\! D_{\phi_2}(\bm{z}_l)\|^2\right]\!+\!D_{KL}(q_{\phi_1}(\cdot |\bm x)\|p(\cdot)),
\end{align}
as a loss function to be is minimized during training. With deterministic encoding, i.e., $q_{\phi_1}(\bm z_l|\bm x)=\delta(\bm z_l-E_{\phi_1}(\bm x))$, then the reconstruction term and the KL divergence term\footnote{The differential entropy of the degenerate distribution $\delta(\bm z_l-E_{\phi_1}(\bm x))$ is mathematically undefined (diverges to $-\infty$), but it is omitted here because it does not depend on the model parameters $\phi_1$ or $\phi_2$. This simplification aligns with standard practice in deterministic encoder settings (e.g., regularized AEs), where only the prior-dependent term $-\ln p(E_{\phi_1}(\bm x))$ contributes to optimization gradients.} in the loss \eqref{vae_obj}  simplify to
\begin{align*}
\mathbb{E}_{q_{\phi_1}\!(\bm z_l|\bm x)}[\|\bm{x}\! -\! D_{\phi_2}(\bm{z}_l)\|^2]&=\|\bm x\!-\!D_{\phi_2}(E_{\phi_1}(\bm x))\|^2\quad \text{and}\\
D_{KL}(q_{\phi_1}(\bm z_l|\bm x)\|p(\bm{z}_l))&\propto -\ln p(E_{\phi_1}(\bm x)).
\end{align*}
Further, by modeling the scaled reconstruction error $z_e=\|\bm x- \bm{x}'\|^2/d_0$ with an exponential-type likelihood, $p(z_e|\bm z_l)=d_0e^{-d_0z_e}$, such that $\log p(z_e|\bm z_l)=c_1-\|\bm x-D_{\phi_2}(E_{\phi_1}(\bm x))\|^2$, where $\bm{x}'= D_{\phi_2}(E_{\phi_1}(\bm x))$, the simplified loss is equal to the negative log-likelihood of the joint distribution $p(\bm z_l,z_e) = p(z_e|\bm z_l)p(\bm z_l)$,
\begin{align}\label{lsa_joint}
&\|\bm x \! -\! \bm x' \|^2\!-\!\ln p(E_{\phi_1}(\bm x)) =c_1 -\ln p(\bm z_l,z_e).
\end{align}
This interpretation provides a theoretical justification for learning a joint distribution over latent feature and reconstruction error for the purpose of AD.  In LSA \citep{abati2019latent},  the prior distribution $p(\bm z_l)$ is replaced with an autoregressive model $p_r$, resulting in the modified training loss,
\begin{align}\label{lsa_obj}
\|\bm x-D_{\phi_2}(E_{\phi_1}(\bm x))\|^2-\beta\ln p_r(E_{\phi_1}(\bm x)),
\end{align}
where the above formula coincides with a deterministic encoder for the $\beta$-VAE \citep{higgins2017beta}, which generalizes the standard VAE by introducing a hyperparameter $\beta$ to scale the weight of the KL divergence term. During inference, the anomaly score is \eqref{lsa_obj} with $\beta=1$. In summary, the anomaly score in LSA considers both the reconstruction error and the distribution of latent features, which corresponds to the ELBO of the data distribution when the VAE encoder is deterministic and the prior is modeled by a probabilistic model.

Similarly, to capture angular errors measured by the cosine dissimilarity of the reconstruction, the decoder output in a VAE can be modeled as a von Mises-Fisher distribution,
\begin{align*}
q_{\phi_2}(\bm x|\bm z_l)&=C_d(\kappa)\exp\left(\kappa\frac{D_{\phi_2}(\bm z_l)^T\bm x}{\|D_{\phi_2}(\bm z_l)\|\|\bm x\|}\right),\\
 \ln q_{\phi_2}(\bm x|\bm z_l)&=c_2-\frac{1}{2}\left(1-\frac{D_{\phi_2}(\bm z_l)^T\bm x}{\|D_{\phi_2}(\bm z_l)\|\|\bm x\|}\right),
\end{align*}
where $C_d(\kappa)=\exp(c_2)$ denotes the normalizing constant and $\kappa=\frac{1}{2}$ in the second expression. Under the same modeling assumptions as before, we could define a new anomaly score as
$$\frac{1}{2}\left(1-\frac{\bm x^T\bm x'}{\|\bm x\|\|\bm x'\|}\right)-\ln p(E_{\phi_1}(\bm x)),$$
which can be interpreted as the negative log-likelihood of the joint distribution $-\ln p(\bm z_l, z_c)$ , where $z_c=\frac{1}{2}\left(1-\frac{\bm x^T\bm x'}{\|\bm x\|\|\bm x'\|}\right)$ represents the cosine dissimilarity between the input $\bm x$ and the reconstructed input $\bm x'$.

This connection motivates modeling the joint distribution of the latent feature $\bm z_l$ with reconstruction error $z_e$ and cosine dissimilarity $z_c$. Accordingly, we construct the composite feature $\bm z=(\bm z_l,z_e,z_c)$ and perform joint density estimation without assuming a specific parametric form.

\subsection{Noise-Contrastive Estimation (NCE)}
We adopt noise-contrastive estimation (NCE) \citep{gutmann2010noise,gutmann2012noise} to train a neural network to produce an estimate of the probability density function $p_d$ of the composite features $\bm{z}$ for nomral data. The fundamental concept behind NCE is to model an unnormalized density function by contrasting it with an auxiliary noise distribution, which is intentionally designed to be tractable for both evaluation and sampling purposes. Given the data distribution $p_d$ and the noise distribution $p_n$,  we define the conditional distributions of $\bm{u}$, which is either data or noise, as
$p_{\bm u|y}(\bm{u}|y)=\begin{cases}p_d(\bm{u}),& y=1\\p_n(\bm{u}),& y=0\end{cases}$, where $y\in\{0,1\}$. Then, the model distribution $p^{\bm\theta}_{\bm z}$ is indirectly fit to the data distribution $p_d$ using the maximum likelihood  estimate of $p^{\bm\theta}_{y|\bm u}$ as
\begin{align}\label{general_nce2}
&\quad\max_{\bm\theta}  \mathbb{E}[\ln p^{\bm\theta}_{y|\bm u}(y|\bm{u})]
=\max_{\bm\theta} \mathbb{E}_{\bm{z}\sim p_d}[\ln p^{\bm\theta}_{y|\bm z}(1|\bm{z})]+\nu \mathbb{E}_{\bm{v}\sim p_n}[\ln p^{\bm\theta}_{y|\bm z}(0|\bm{v})],
\end{align}
where $\nu$ denotes $\frac{\mathrm{Pr}(y=0)}{\mathrm{Pr}(y=1)}$. The posterior probability $p^{\bm\theta}_{y|\bm u}$ is modeling using logistic regression
\begin{align*}
p^{\bm\theta}_{y|\bm u}(y=1|\bm{u})&=\frac{p^{\bm\theta}_{\bm z}(\bm{u})}{p^{\bm\theta}_{\bm z}(\bm{u})+\nu p_n(\bm{u})}\\
&=\sigma(\ln p^{\bm\theta}_{\bm z}(\bm{u})-\ln\nu p_n(\bm{u}))=\sigma(T_{\bm\theta}(\bm{u})),
\end{align*}
where $\sigma(x)=\frac{1}{1+e^{-x}}$ is the sigmoid function and $T_{\bm\theta}$ is a neural network model of the the log-odds
$\ln p^{\bm\theta}_{\bm z}(\bm{u})-\ln\nu p_n(\bm{u})$. Substituting the expressions of
$p^{\bm\theta}_{y|\bm u}(1|\bm{u})$ and $p^{\bm\theta}_{y|\bm u}(0|\bm{u})=1-p^{\bm\theta}_{y|\bm u}(1|\bm{u})$ in terms of $T_\theta$
into \eqref{general_nce2}, and negating the objective
we obtain the loss function
\begin{align}\label{nce}
&\quad\mathcal{L}_{\mathrm{NCE}}(\bm\theta)
=-\underset{{\bm{z}\sim p_d}}{\mathbb{E}}\left[\ln\sigma\left(T_{\bm\theta}(\bm{z})\right)\right]-\nu\underset{\bm{v}\sim p_n}{\mathbb{E}}\left[\ln(1-\sigma\left(T_{\bm\theta}(\bm{v}))\right)\right].
\end{align}
Substituting $\bm\theta^*$, the minimizer of $\mathcal{L}_{\mathrm{NCE}}(\bm\theta)$, into the log-odds and rearranging the terms yields
$\ln p^{\bm\theta^*}_{\bm z}(\bm z) = T_{\bm\theta^*}(\bm z)+\ln\nu p_n(\bm z) = -S_C(\bm z)$, where $S_C$ is the anomaly score on the composite features. When the model is sufficiently powerful, the optimal model $p^{\bm\theta^*}_{y|\bm{u}}$ will match $p_{y|\bm u}$, implying that $p^{\bm\theta^*}_{\bm z} \equiv p_d$ and $S_C(\bm z) \equiv - \ln p_d(\bm z)$.
\subsection{Adapting NCE for AD}
Selecting an appropriate noise distribution $p_n(\bm{z})$ is crucial for the success of NCE.  As discussed in \citep{gutmann2010noise}, NCE performs optimally when the noise distribution $p_n$ closely resembles the composite feature distribution $p_d$. Following this principle, we iteratively optimize the noise distribution during NCE training.

\subsubsection{Optimizing Noise Distribution}
In NCE, the noise distribution $p_n$ is often chosen to be Gaussian $\mathcal{N}(\hat{\bm\mu}_{\bm z},\hat{\bm\Sigma}_{\bm z})$, where $\hat{\bm\mu}_{\bm z}$ and $\hat{\bm\Sigma}_{\bm z}$ are the sample mean and variance derived from the training dataset, respectively.\footnote{For large training datasets, these two estimators are not feasible because the entire dataset cannot be processed by the compression network simultaneously. To address this limitation, we can iteratively estimate the mean and covariance of the sample in batches as detailed in Appendix \ref{iterative_mean}.}  We create a refined the noise distribution for NCE through the parametrization $p_{n\bm{K}}=\mathcal{N}(\hat{\bm\mu}_{\bm z},\bm{K}^T\hat{\bm\Sigma}_{\bm z}\bm{K})$, where $\bm{K}$ represents a parameter matrix. Subsequently, reparameterization trick makes it feasible to optimize $\bm{K}$ to maximize the NCE loss through backpropagation. We first draw a sample $\bm{z}\sim\mathcal{N}(\hat{\bm\mu}_{\bm z}, \hat{\bm\Sigma}_{\bm z})$ and then use the affine function $L_{\bm K}(\bm z)=\bm{K}(\bm{z}-\hat{\bm\mu}_{\bm z})+\hat{\bm\mu}_{\bm z}$ to draw from the intended Gaussian $\mathcal{N}(\hat{\bm\mu}_{\bm z}, \bm{K}^T\hat{\bm\Sigma}\bm{K})$. However, the naive maximization of the NCE loss in terms of $\bm{K}$ is equivalent to minimizing $\mathbb{E}_{\bm{u}\sim p_n}[\ln(1-\sigma(T_{\bm\theta}(L_{\bm K}(\bm u))))]$, since the first term in \eqref{nce} does not depend on noise. Instead, following similar work for GAN training \citep{che2016mode}, the first term in the NCE loss can be incorporated into the optimization as
\begin{align}\label{adnce}
\min_{\bm{K}}\ &\ \mathbb{E}_{\bm{u}\sim p_d}\left[\ln\sigma\left(T_{\bm\theta}\left(L_{\bm K}\left(\textbf{sg}\left(L_{\bm K}^{-1}\left(\bm{u}\right)\right)\right)\right)\right)\right]
+\nu\mathbb{E}_{\bm{u}\sim p_n}\left[\ln\left(1-\sigma\left(T_{\bm\theta}\left(L_{\bm K}\left(\bm{u}\right)\right)\right)\right)\right],
\end{align}
where $L^{-1}_{\bm K}(\bm z)=\bm{K}^{-1}(\bm{z}-\hat{\bm\mu}_{\bm z})+\hat{\bm\mu}_{\bm z}$ and $\textbf{sg}(\cdot)$ is the stop gradient operation.

\subsubsection{Augmented Reconstruction Features}
Well-chosen data augmentation techniques generally enhance model performance. However, in AD augmentations that preserve normality require domain knowledge. We propose to augment normal data by adjusting the reconstruction features alone, without modifying the input or latent representations.\footnote{Data augmentation of the input improves model performance \citep{golan2018deep}, but it is beyond the scope of this paper, as our model and other baselines do not incorporate it.} We achieve this by generating additional normal points by replacing reconstruction features with artificially lower values while maintaining the latent representations. The goal is to bias the estimation network such that artificially low reconstruction features, which may be observed by chance in points the noise distribution, are deemed normal.

Specifically, we create artificial normal points by defining $\bm z = (\bm z_l, z_e', z_c')$, where $\bm z_l$ is the latent feature of normal data, and $z_e'\sim p_{t_1}$ and $z_c'\sim p_{t_2}$ are independently drawn from the truncated normal distributions. Next, we explain how the parameters of the truncated normal distributions $p_{t_1}$ and $p_{t_2}$ are defined to ensure that each marginals of the augmented data distribution have a density that is higher than the noise distribution over low reconstruction errors/dissimilarities.

Given that the reconstruction feature exhibits a skewed unimodal form, we assume it follows a log-normal distribution, $\ln z\sim\mathcal{N}(\mu, \sigma)$, where $z\sim p_t$ is either $z_e\sim p_{t_1}$ or $z_c\sim p_{t_2}$. Then, its density function is
$p_0(z)=\frac{1}{z\sigma\sqrt{2\pi}}\exp\left[-\frac{(\ln z-\mu)^2}{2\sigma^2}\right]$.
Its mode $m_z$ can be computed using its mean $\mu_z$ and variance $\sigma_z$ as
$m_z=\mu_z(\frac{\sigma_z^2}{\mu_z^2}+1)^{-\frac{3}{2}}$, as shown in Appendix \ref{mode_dev}. We estimate the distribution mean $\mu_z$, distribution variance $\sigma_z$, and distribution mode $m_z$ using the training dataset. These estimates are then used to define the truncated normal distribution $p_t$ defined on $[0,m_z]$. With an equal mixture of normal and artificial normal points, the augmented density function becomes
\begin{align}
p_m(\bm{z})=\frac{1}{2}p_d(\bm{z})+\frac{1}{2}p_l(\bm{z}_l)p_{t_1}(z_e)p_{t_2}(z_c),
\end{align}
where $p_l$ represents the marginal distribution of $p_d$ with respect to the latent feature $\bm{z}_l$, and $p_{t_1}$ and $p_{t_2}$ are the truncated normal distributions for the reconstruction features $z_e$ and $z_c$, respectively.  $p_m$ is substituted for $p_d$ in \eqref{nce} during the optimization of the estimation network. The following proposition (proof in Appendix~\ref{proposition_proof}) provides insight into the effect of the data augmentation strategy. Specifically, it shows that the augmented normal density $p_m$ has reconstruction-feature marginals no smaller than those of the noise distribution, which helps reduce the risk of normal samples being misclassified as anomalies.
\begin{proposition}\label{prop}
The density of the marginal distribution of the reconstruction feature in $p_m$ is no less than the density of the corresponding marginal distribution of the noise distribution $p_n$ over the interval $[0,m_z]$.
\end{proposition}

\subsubsection{Implementation}
Taking into account the parameterized noise distribution, we adopt an alternating optimization strategy with the batch loss for the estimation network $T_{\bm\theta}$ as \begin{align}\notag
-\frac{1}{M}\sum_{i=1}^M\ln\sigma(T_{\bm\theta}(\bm{z}_i)) -\frac{\nu}{N}\sum_{i=1}^N\ln(1-\sigma(T_{\bm\theta}(L_{\bm K}(\bm{v}_i)))),
\end{align}
where $\nu=\frac{N}{M}$ is the noise-sample ratio, $\bm{z}_1,\ldots,\bm{z}_M$ is sampled from the augmented data $\bm z_i \sim p_m$ and $\bm{v}_1,\ldots, \bm{v}_N$ is sampled from the noise distribution $\bm v_i \sim p_n$. The corresponding batch loss version of \eqref{adnce} is used to update $\bm{K}$.
Furthermore, we constrain $\bm{K}$ to be a diagonal matrix with diagonal elements equal to or greater than $1$. This constraint enhances training stability (noise variance can only grow) and facilitates a more efficient computation of the inverse of the affine transformation $L_{\bm K}$. In practice, the constraint is enforced via softplus $K_{jj}=1+\log(1+\exp(\psi_j)),\bm{\psi}\in\mathbb{R}^d$. AdamW is used for both sets of parameters $\bm{\theta}$ and $\bm{\psi}$.

\section{Experiments}
We evaluate CANCE through two sets of experiments. First, we conduct ablation studies on MNIST \citep{lecun2010mnist}, Fashion-MNIST (F-MNIST) \citep{xiao2017fashion}, CIFAR-10 \citep{krizhevsky2010cifar}, and MNIST-C \citep{mu2019mnist} to examine the contributions of composite features and augmented reconstruction features. Second, we benchmark CANCE against existing AD methods on ADBench \citep{han2022adbench}, a large-scale anomaly detection benchmark comprising 57 datasets. Throughout all experiments, we split the training data into training and validation sets using an 80/20 split for model training and selection, respectively, and set the NCE noise-to-data ratio to $\nu=8$. We use AUROC as the evaluation metric throughout all experiments.

\subsection{Ablation Study}

Following prior work \citep{abati2019latent}, we construct \textbf{unimodal} AD tasks from labeled image datasets by treating one class as normal and all remaining classes as anomalies. Only normal samples from the training set are used for training and model selection, while the entire test set is used for evaluation. For \textbf{multimodal} tasks, two or more classes are treated as normal, with the same training and evaluation protocol. For MNIST-C, we follow \citep{lee2023semi}: the entire MNIST training set is used for training, while at test time the original MNIST images are treated as normal and their corrupted counterparts in MNIST-C \citep{mu2019mnist} as anomalies. Each experiment is repeated five times, and we report the mean and standard deviation, except for MNIST-C, for which we follow \citep{lee2023semi} and report the mean over 30 trials.

We evaluate the contributions of individual components of CANCE using four variants: \textbf{Error}, which uses only the AE reconstruction error as the anomaly score; \textbf{LatNCE}, which applies NCE to the AE latent features; \textbf{CNCE}, which applies NCE to the composite features; and \textbf{CANCE}, which applies NCE to the composite features with augmented reconstruction features.

Table~\ref{ablation-summary} summarizes the results on unimodal MNIST, F-MNIST, and CIFAR-10 and multimodal settings, including MNIST-C,
reporting the mean performance across individual cases within each dataset. CNCE outperforms Error in almost all settings,
demonstrating the benefit of combining latent and reconstruction features. CNCE also consistently outperforms LatNCE,
further supporting the importance of incorporating reconstruction information into the density estimation.
Appendix~\ref{ablation} provides full unimodal and multimodal results in Tables~\ref{ablation-unimode} and \ref{ablation-multimodal}, respectively,
along with experimental details. As shown in the full results, CANCE achieves equal or better mean performance than CNCE in 17 of the 20 unimodal settings
and across all seven multimodal settings, demonstrating the additional benefit of augmented reconstruction features.

\begin{table}[tbh]
\centering
\caption{AUROC for CANCE ablations}
\label{ablation-summary}
\begin{tabular}{lccccc}
\hline
Data &Error & LatNCE & CNCE & CANCE\\
\hline
\multicolumn{3}{l}{Unimodal}\\
MNIST & $95.1$ & $84.1$ & $97.0$ & $97.7$ \\
F-MNIST & $90.4$ & $89.6$ & $92.2$ & $92.5$ \\
CIFAR-10 &$54.1$ & $65.1$ & $65.3$ & $65.4$ \\
\hline
\multicolumn{3}{l}{Multimodal}\\
MNIST &$94.6$ & $87.6$ & $97.3$ & $97.6$\\
F-MNIST & $86.7$ & $89.0$ & $90.6$ & $90.8$  \\
CIFAR-10 & $53.0$ & $60.5$ & $60.8$ & $60.9$ \\
MNIST-C &$89.7$ & $78.4$ & $91.3$ & $92.2$ \\
\hline
\end{tabular}
\end{table}

\subsection{Benchmark Comparison on ADBench}
We compare CANCE with DTE \citep{livernoche2024diffusion} and other anomaly detection baselines on ADBench \citep{han2022adbench},
using the same 57 datasets and following the dataset preparation protocol of DTE for a consistent comparison.

As Table~\ref{tab:overall_adbench} shows, CANCE achieves a mean AUROC of 81.23\%, which is comparable to the state-of-the-art DTE-NP method (81.59\%)
while outperforming all other baselines on average. Furthermore, CANCE attains the highest performance on 22 of the 57 datasets, including ties with other
top-performing methods. Detailed dataset-wise results and experimental configurations are provided in Appendix~\ref{sec:adbench_detailed}.

\begin{table}[tbh]
\centering
\caption{Performance comparison across 57 datasets in terms of average AUROC and best count.
The best count indicates the number of datasets on which each method achieves the highest AUROC.}
\label{tab:overall_adbench}
\begin{tabular}{lcc}
\hline
Method & Mean AUROC (\%) & Best Count \\
\hline
DAGMM  & 65.05 & 1  \\
DSVDD  & 67.89 & 0  \\
DROCC  & 60.30 & 0  \\
GOAD   & 68.82 & 3  \\
DDPM   & 77.92 & 3  \\
DTE-NP & 81.59 & 19 \\
DTE-IG & 77.85 & 6  \\
DTE-C  & 80.34 & 7  \\
CANCE  & 81.23 & 22 \\
\hline
\end{tabular}
\end{table}

\section{Conclusion}
In this work, we propose an unsupervised AD framework, CANCE, based on joint density modeling of composite AE features. Motivated by the observation that latent features extracted by AEs alone can be insufficient for AD, we complement the latent representation with reconstruction features and NCE to estimate their joint density. We further establish a connection between joint density modeling of the composite features and the $\beta$-VAE objective, and introduce an augmentation strategy that encourages the estimated density to decrease as reconstruction errors increase. Experiments across multiple benchmark datasets show that CANCE achieves performance comparable to state-of-the-art AD methods while remaining relatively straightforward to implement compared with more complex deep probabilistic models. Moreover, CANCE is not restricted to representations learned directly from raw inputs: AEs can also be applied to pretrained embeddings, making the approach potentially applicable to a broad range of pretrained representation models and domains.

% \subsubsection*{References}

% References follow the acknowledgements. Use an unnumbered third level
% heading for the references section.  Please use the same font
% size for references as for the body of the paper---remember that
% references do not count against your page length total.

% \begin{thebibliography}{}
% \setlength{\itemindent}{-\leftmargin}
% \makeatletter\renewcommand{\@biblabel}[1]{}\makeatother
% \bibitem{} J.~Alspector, B.~Gupta, and R.~B.~Allen (1989).
%     \newblock Performance of a stochastic learning microchip.
%     \newblock In D. S. Touretzky (ed.),
%     \textit{Advances in Neural Information Processing Systems 1}, 748--760.
%     San Mateo, Calif.: Morgan Kaufmann.

% \bibitem{} F.~Rosenblatt (1962).
%     \newblock \textit{Principles of Neurodynamics.}
%     \newblock Washington, D.C.: Spartan Books.

% \bibitem{} G.~Tesauro (1989).
%     \newblock Neurogammon wins computer Olympiad.
%     \newblock \textit{Neural Computation} \textbf{1}(3):321--323.
% \end{thebibliography}

\bibliography{references}

\clearpage
\appendix
\thispagestyle{empty}

% Supplementary material: To improve readability, you must use a single-column format for the supplementary material.
\onecolumn
\aistatstitle{Supplementary Materials}
\section{Proof for Proposition 1}\label{proof-motivation}
Proposition 1. Let $X$ be a random variable with $0<\operatorname{Var}(X)<\infty$, and let $Y=(Y_1,Y_2)^T$, where $Y_i=X+\epsilon_i, \epsilon_i\sim\mathcal{N}(0,\sigma^2)$, and $X,\epsilon_1,\epsilon_2$ are mutually independent. For a one-dimensional encoder $Z=h(a^TY+b)$ with injective $h$ and an arbitrary nonlinear decoder $D$, the minimum reconstruction error
$$
\min_{a,b,h,D}
\mathbb{E}\!\left[
\left\|Y-D\!\left(h(a^TY+b)\right)\right\|^2
\right]
$$
is attained by the latent representation
$Z^\ast=h\left(\frac{Y_1+Y_2}{\sqrt{2}}\right)$,
up to an invertible reparameterization of $Z^\ast$, with optimal reconstruction
$\hat{Y}^\ast=\left(
\frac{Y_1+Y_2}{2},
\frac{Y_1+Y_2}{2}
\right)^T
$.\footnote{
The optimal one-dimensional encoder preserves the signal direction
$\frac{Y_1 + Y_2}{\sqrt{2}}$, whereas the orthogonal direction $\frac{Y_1 - Y_2}{\sqrt{2}}$ is discarded from the latent representation and appears only in the reconstruction residual.}

Proof. Step 1. For a fixed encoder, the optimal decoder is the conditional mean.

For a fixed encoder $Z=h(a^T Y+b)$, by the orthogonality property of
conditional expectation,
\[
\mathbb{E}\!\left[\|Y-D(Z)\|^2\right]
=
\mathbb{E}\!\left[\|Y-\mathbb{E}[Y\mid Z]\|^2\right]
+
\mathbb{E}\!\left[\|\mathbb{E}[Y\mid Z]-D(Z)\|^2\right].
\]
Therefore, the optimal decoder is
\[
D^\ast(Z)=\mathbb{E}[Y\mid Z],
\]
with minimum reconstruction error
\[
\min_D \mathbb{E}\!\left[\|Y-D(Z)\|^2\right]
=
\mathbb{E}\!\left[\|Y-\mathbb{E}[Y\mid Z]\|^2\right].
\]

Then,
\begin{eqnarray*}
  R (Z) & = & \min_D \mathbb{E} [\| Y - D(Z) \|^2]\\
  & = & \mathbb{E} [\| Y -\mathbb{E} [Y|Z] \|^2]\\
  & = & \mathbb{E} [(Y -\mathbb{E} [Y|Z])^T (Y -\mathbb{E} [Y|Z])]\\
  & = & \mathbb{E} [\operatorname{tr} ((Y -\mathbb{E} [Y|Z])^T (Y -\mathbb{E}
  [Y|Z]))]\\
  & = & \mathbb{E} [\operatorname{tr} ((Y -\mathbb{E} [Y|Z]) (Y -\mathbb{E}
  [Y|Z])^T)]\\
  & = & \operatorname{tr} (\mathbb{E} [(Y -\mathbb{E} [Y|Z]) (Y -\mathbb{E}
  [Y|Z])^T])\\
  & = & \operatorname{tr} (\mathbb{E}_Z [\mathbb{E} [(Y -\mathbb{E} [Y|Z]) (Y
  -\mathbb{E} [Y|Z])^T |Z]])\\
  & = & \operatorname{tr} (\mathbb{E}_Z [\operatorname{Cov} (Y|Z)])\\
  & = & \mathbb{E}_Z [\operatorname{tr}(\operatorname{Cov} (Y|Z))] .
\end{eqnarray*}

Step 2. Invariance under injective transformation

Let $$T=a^T Y+b,$$ so that \(Z=h(T)\). Since \(h\) is injective, \(T\) can be recovered from \(Z\), and hence \(Z\) and \(T\) generate the same information. Therefore,
$$\mathbb{E}[Y\mid Z]=\mathbb{E}[Y\mid T],$$
and
$$\operatorname{Cov}(Y\mid Z)=\operatorname{Cov}(Y\mid T).$$
It follows that
$$
R(Z)=
\mathbb{E}\!\left[
\operatorname{tr}\!\left(\operatorname{Cov}(Y\mid Z)\right)
\right]
=\mathbb{E}\!\left[
\operatorname{tr}\!\left(\operatorname{Cov}(Y\mid T)\right)
\right]
=R(T).
$$
Thus, the injective transformation $h$ does not affect the optimal reconstruction error.

Step 3. Remove bias and scaling

By the same argument as in Step 2, any invertible affine transformation of \(T\) preserves the optimal reconstruction error. Since shifts and nonzero scalings are invertible, without loss of generality, we set
$$
\|a\|_2=1,\qquad b=0.
$$
Thus,
$$
T=a^T Y, \quad\|a\|_2=1.
$$

Step 4. Introduce the orthogonal direction

Let $a_{\perp} = \left[ \begin{array}{l}
  - a_2\\
  a_1
\end{array} \right]$. Then $\| a_{\perp} \| = 1$ and $a^Ta_{\perp}= 0$, so $\{ a, a_{\perp} \}$ forms an orthonormal basis for $\mathbb{R}^2$. Hence,
\begin{eqnarray*}
  Y & = & I Y\\
  & = & (a a^T + a_{\perp}a_{\perp}^T) Y\\
  & = & a (a^T Y) + a_{\perp}(a_{\perp}^T Y)\\
  & = & a T + a_{\perp} W
\end{eqnarray*}
where $W = a_{\perp}^T Y = - a_2 Y_1 + a_1 Y_2$. Since $T$ is available to the decoder, $aT$ can be reconstructed exactly, so only the $W$-component contributes to the reconstruction error.

Step 5. Reduce the reconstruction error to $W$

For any decoder $D(Z)$, because $\{ a, a_{\perp} \}$ forms an orthonormal basis for
$\mathbb{R}^2$,
\begin{eqnarray*}
D(Z) & = & I g (Z)\\
& = & (a a^T + a_{\perp}a_{\perp}^T)D(Z)\\
& = & a (a^T D(Z)) + a_{\perp} (a_{\perp}^T D(Z))\\
& = & a D_1(Z) + a_{\perp}D_2 (Z)
\end{eqnarray*}
where $D_1 (Z) = a^T D(Z)$ and $D_2 (Z) = a_{\perp}^T D(Z)$. Therefore
\begin{eqnarray*}
\| Y - D(Z) \|^2 & = & \| a T + a_{\perp} W - a D_1 (Z) + a_{\perp} D_2 (Z) \|^2\\
& = & \| a [T - D_1 (Z)] + a_{\perp} [W - D_2 (Z)] \|^2\\
& = & [T - D_1 (Z)]^2 + [W - D_2 (Z)]^2 .
\end{eqnarray*}
Note that $Z = h(T)$ and $h$ is injective. So the optimal decoder $D_1^{\ast}
= h^{- 1}$ can reconstruct $T$ exactly. Repeating the argument in Step 1 for the scalar $W$,
$$
\begin{aligned}
R(Z)
&=\mathbb{E}\!\left[(W-\mathbb{E}[W\mid Z])^2\right]\\
&=\mathbb{E}\!\left[\operatorname{Var}(W\mid Z)\right].
\end{aligned}
$$

Since $Z=h(T)$ with injective $h$, $Z$ and $T$ contain the same information. Hence
$$
R(Z)
=\mathbb{E}\!\left[\operatorname{Var}(W\mid T)\right].
$$

Step 6. Express $T$ and $W$ in terms of signal and noise

Using \(Y_i=X+\epsilon_i\),
$$
\begin{aligned}
T&=a_1Y_1+a_2Y_2\\
&=(a_1+a_2)X+a_1\epsilon_1+a_2\epsilon_2.
\end{aligned}
$$

Let $s=a_1+a_2$ and $N=a_1\epsilon_1+a_2\epsilon_2$. Then $T=sX+N$. Since \(\|a\|_2=1\), $N\sim\mathcal N(0,\sigma^2)$.
Similarly,
$$
\begin{aligned}
W
&=-a_2Y_1+a_1Y_2\\
&=(a_1-a_2)X-a_2\epsilon_1+a_1\epsilon_2.
\end{aligned}
$$
Let
$q=a_1-a_2$ and $M=-a_2\epsilon_1+a_1\epsilon_2$. Then $W=qX+M$, where $
M\sim\mathcal N(0,\sigma^2)$.

% Step 7. $M$ is independent of $T$
% \begin{eqnarray*}
%   \mathbb{Cov} (N, M) & = & \mathbb{Cov} (a_1 E_1 + a_2 E_2, - a_2 E_1 + a_1
%   E_2)\\
%   & = & - a_1 a_2 \sigma^2 + a_1 a_2 \sigma^2\\
%   & = & 0
% \end{eqnarray*}
% Since $M, N$ are jointly Gaussian, $N \perp M$. Furthermore, since $X$ is
% independent of $(E_1, E_2)$, and hence independent of $(N, M)$. Combing $X
% \perp (N, M)$ and $N \perp M$, we establish the joint independence $M \perp (X,
% N)$. Since $T = s X + N$, we obtain
% \[ M \perp T. \]
% Consequently,
% \[ \mathbb{E} [M|T] = 0 \]
% and
% \[ \mathbb{Var} (M|T) = \sigma^2 . \]

Step 7. Show that $M$ is independent of $T$

Since
$$
\begin{aligned}
\operatorname{Cov}(N,M)
&=\operatorname{Cov}
(a_1\epsilon_1+a_2\epsilon_2,-a_2\epsilon_1+a_1\epsilon_2)\\
&=-a_1a_2\sigma^2+a_1a_2\sigma^2\\
&=0,
\end{aligned}
$$
and $N$ and $M$ are jointly Gaussian, we have $N\perp M$. Moreover, since $X\perp(\epsilon_1,\epsilon_2)$, we have $X\perp(N,M)$. Together with $N\perp M$, it follows that $X, N$, and $M$ are mutually independent, and hence $M\perp(X,N)$. Since $T=sX+N$ is a function of $(X,N)$, $M\perp T$. Consequently,
$$
\mathbb{E}[M\mid T]=0,
\qquad
\operatorname{Var}(M\mid T)=\sigma^2.
$$

Step 8. Compute the conditional variance of $W$

Recall that $W=qX+M$. Therefore,
$$
\operatorname{Var}(W\mid T)
=q^2\operatorname{Var}(X\mid T)
+\operatorname{Var}(M\mid T)
+2q\operatorname{Cov}(X,M\mid T).
$$
By Step 7, $M\perp(X,N)$. Since $T=sX+N$ is a function of $(X,N)$, we have $M\perp(X,T)$, which implies $M\perp X\mid T$. Therefore,
$$
\operatorname{Var}(M\mid T)=\sigma^2,
\qquad
\operatorname{Cov}(X,M\mid T)=0.
$$

It follows that
$$
\operatorname{Var}(W\mid T)=(a_1-a_2)^2\operatorname{Var}(X\mid T)+\sigma^2.
$$

Therefore,
\begin{eqnarray*}
R (Z) & = & R (T)\\
& = & \mathbb{E} [\operatorname{Var} (W|T)]\\
& = & \mathbb{E} [(a_1 - a_2)^2 \operatorname{Var} (X|T) + \sigma^2]\\
& = & (a_1 - a_2)^2 \mathbb{E} [\operatorname{Var} (X|T)] + \sigma^2\\
& \geq & \sigma^2 .
\end{eqnarray*}

Step 9. Show that $\mathbb{E} [\operatorname{Var} (X|T)] > 0$.

Recall that $T=sX+N$, where $s=a_1+a_2, 0<\operatorname{Var}(X)<\infty, N\sim\mathcal{N}(0,\sigma^2), \sigma>0$, and $X\perp N$. If $s=0$, then $T=N\perp X$, so
$$
\mathbb{E}[\operatorname{Var}(X\mid T)]
=\operatorname{Var}(X)>0.
$$
It remains to consider \(s\neq0\).
% By the law of total variance,
% $$
% \operatorname{Var}(X)
% =\mathbb{E}[\operatorname{Var}(X\mid T)]
% +\operatorname{Var}(\mathbb{E}[X\mid T]).
% $$
Suppose, for contradiction, that $\mathbb{E} [\operatorname{Var} (X|T)] = 0$. Since $\operatorname{Var}(X\mid T)\geq 0$, it follows that
\[ \operatorname{Var} (X|T) = 0\quad\text{a.s.}, \]
which implies
\[ X =\mathbb{E} [X|T]\quad\text{a.s.} \]
Thus, $X$ can be recovered from $T$ almost surely. However,
$$ T=sX+N,$$
where $N\sim\mathcal N(0,\sigma^2)$ is independent of $X$ with $\sigma^2>0$. Since the Gaussian noise has full support, the conditional distributions $T\mid X=x$ overlap for distinct values of $x$. Hence, $X$ cannot be recovered from $T$ almost surely, yielding a contradiction. Hence
$$ \mathbb{E} [\operatorname{Var} (X|T)] > 0.$$

Step 10. Optimal latent representation and reconstruction.

Since $\mathbb{E}[\operatorname{Var}(X\mid T)]>0$,
$$
R(Z)
=
(a_1-a_2)^2\mathbb{E}[\operatorname{Var}(X\mid T)]
+\sigma^2
$$
is minimized if and only if $a_1=a_2$. Given the constraint $a_1^2+a_2^2=1$, the optimal projection vector is
$$
a^\ast=\pm\frac{1}{\sqrt{2}}(1,1)^T,
$$
yielding the optimal latent representation
$$
Z^\ast=h\left(\pm\frac{Y_1+Y_2}{\sqrt{2}}\right),
$$
with minimum reconstruction error
$$
R(Z^\ast)=\sigma^2.
$$
By Step 1,
$$D^\ast(Z^\ast)=\mathbb{E}[Y\mid Z^\ast].$$
Since $h$ is injective, conditioning on $Z^\ast$ is equivalent to conditioning on
$$
T^\ast=\pm\frac{Y_1+Y_2}{\sqrt{2}}.
$$
Using the orthogonal decomposition from Step 4,
$$
Y=a^\ast T^\ast+a_\perp^\ast W^\ast.
$$
At the optimum, $q=a_1-a_2=0$. Hence, from $W=qX+M$,
$$
W^\ast=M,
$$
and by Step 7,
$$
\mathbb{E}[W^\ast\mid T^\ast]=0.
$$
Therefore, the optimal reconstruction is
$$
\begin{aligned}
D^\ast(Z^\ast)
=\mathbb{E}[Y\mid T^\ast]
=a^\ast T^\ast+a_\perp^\ast\mathbb{E}[W^\ast\mid T^\ast]=a^\ast T^\ast\
=\left(\frac{Y_1+Y_2}{2},\frac{Y_1+Y_2}{2}\right)^T.
\end{aligned}
$$

\section{Mean and Variance Derivation}\label{iterative_mean}
By the definition of the sample mean, we have
\begin{align}\label{update_mean}
\hat{\bm\mu}_{\bm z}^{(t+1)}&=\frac{1}{n_t+n_b}\sum_{i=1}^{n_t+n_b}\bm{z}_i\nonumber\\
&=\frac{1}{n_t+n_b}\left(\sum_{i=1}^{n_t}\bm{z}_i+\sum_{i=n_t+1}^{n_t+n_b}\bm{z}_i\right)\nonumber\\
&=\frac{n_t\hat{\bm\mu}_{\bm z}^{(t)}+n_b\hat{\bm\mu}_{\bm z}^{(b)}}{n_t+n_b}
\end{align}
By definition, the sample covariance matrix is as follows:
\begin{align*}
\hat{\bm\Sigma}_{\bm z}^{(t+1)}&=\frac{1}{n_t+n_b}\sum_{i=1}^{n_t+n_b}
\left(\bm{z}_i-\hat{\bm\mu}_{\bm z}^{(t+1)}\right)^T
\left(\bm{z}_i-\hat{\bm\mu}_{\bm z}^{(t+1)}\right)\\
&=\frac{1}{n_t+n_b}\left[\sum_{i=1}^{n_t}
\left(\bm{z}_i-\hat{\bm\mu}_{\bm z}^{(t+1)}\right)^T
\left(\bm{z}_i-\hat{\bm\mu}_{\bm z}^{(t+1)}\right)+\sum_{i=n_t+1}^{n_t+n_b}\left(\bm{z}_i-\hat{\bm\mu}_{\bm z}^{(t+1)}\right)^T
\left(\bm{z}_i-\hat{\bm\mu}_{\bm z}^{(t+1)}\right)\right]
\end{align*}
Consider expanding the first sum term as follows,
\begin{align*}
&\quad \sum_{i=1}^{n_t}\left(\bm{z}_i-\hat{\bm\mu}_{\bm z}^{(t+1)}\right)^T
\left(\bm{z}_i-\hat{\bm\mu}_{\bm z}^{(t+1)}\right)\\
&=\sum_{i=1}^{n_t}\left(\bm{z}_i-\hat{\bm\mu}_{\bm z}^{(t)}+
\hat{\bm\mu}_{\bm z}^{(t)}-\hat{\bm\mu}_{\bm z}^{(t+1)}\right)^T
\left(\bm{z}_i-\hat{\bm\mu}_{\bm z}^{(t)}+\hat{\bm\mu}_{\bm z}^{(t)}-\hat{\bm\mu}_{\bm z}^{(t+1)}\right)\\
&=\sum_{i=1}^{n_t}\left(\bm{z}_i-\hat{\bm\mu}_{\bm z}^{(t)}\right)^T
\left(\bm {z}_i-\hat{\bm\mu}_{\bm z}^{(t)}\right)\quad+\sum_{i=1}^{n_t}
\left(\hat{\bm\mu}_{\bm z}^{(t)}-\hat{\bm\mu}_{\bm z}^{(t+1)}\right)^T\left(\hat{\bm\mu}_{\bm z}^{(t)}-\hat{\bm\mu}_{\bm z}^{(t+1)}\right)\\
&+\sum_{i=1}^{n_t}\left(\bm{z}_i-\hat{\bm\mu}_{\bm z}^{(t)}\right)^T
\left(\hat{\bm\mu}_{\bm z}^{(t)}-\hat{\bm\mu}_{\bm z}^{(t+1)}\right)\sum_{i=1}^{n_t}\left(\hat{\bm\mu}_{\bm z}^{(t)}-\hat{\bm\mu}_{\bm z}^{(t+1)}\right)^T
\left(\bm{z}_i-\hat{\bm\mu}_{\bm z}^{(t)}\right)\\
\end{align*}
By calculation, we see
\begin{align*}
\sum_{i=1}^{n_t}\left(\bm{z}_i-\hat{\bm\mu}_{\bm z}^{(t)}\right)^T
\left(\hat{\bm\mu}_{\bm z}^{(t)}-\hat{\bm\mu}_{\bm z}^{(t+1)}\right)&=\left[\sum_{i=1}^{n_t}\left(\bm{z}_i-\hat{\bm\mu}_{\bm z}^{(t)}\right)^T\right]\left(\hat{\bm\mu}_{\bm z}^{(t)}-\hat{\bm\mu}_{\bm z}^{(t+1)}\right)\\
&=\left[\sum_{i=1}^{n_t}\bm{z}_i-\hat{\bm\mu}_{\bm z}^{(t)}\right]^T
\left(\hat{\bm\mu}_{\bm z}^{(t)}-\hat{\bm\mu}_{\bm z}^{(t+1)}\right)\\
&=\left[\sum_{i=1}^{n_t}\bm{z}_i-\sum_{i=1}^{n_t}\hat{\bm\mu}_{\bm z}^{(t)}\right]^T
\left(\hat{\bm\mu}_{\bm z}^{(t)}-\hat{\bm\mu}_{\bm z}^{(t+1)}\right)\\
&=[n_t\hat{\bm\mu}_{\bm z}^{(t)}-n_t\hat{\bm\mu}_{\bm z}^{(t)}]^T\left(\hat{\bm\mu}_{\bm z}^{(t)}
-\hat{\bm\mu}_{\bm z}^{(t+1)}\right)\\
&=\bm0
\end{align*}
Similarly,
\begin{align*}
\sum_{i=1}^{n_t}\left(\hat{\bm\mu}_{\bm z}^{(t)}-\hat{\bm\mu}_{\bm z}^{(t+1)}\right)^T
\left(\bm{z}_i-\hat{\bm\mu}_{\bm{z}}^{(t)}\right)=\bm0.
\end{align*}
Thus,
\begin{align*}
&\quad\sum_{i=1}^{n_t}\left(\bm{z}_i-\hat{\bm\mu}_{\bm z}^{(t+1)}\right)^T
\left(\bm{z}_i-\hat{\bm\mu}_{\bm z}^{(t+1)}\right)\\
&=n_t\hat{\bm\Sigma}_{\bm z}^{(t)}+\sum_{i=1}^{n_t}
\left(\hat{\bm\mu}_{\bm z}^{(t)}-\hat{\bm\mu}_{\bm z}^{(t+1)}\right)^T
\left(\hat{\bm\mu}_{\bm z}^{(t)}-\hat{\bm\mu}_{\bm z}^{(t+1)}\right)\\
&=n_t\hat{\bm\Sigma}_{\bm z}^{(t)}+n_t\left(\hat{\bm\mu}_{\bm z}^{(t)}
-\hat{\bm\mu}_{\bm z}^{(t+1)}\right)^T
\left(\hat{\bm\mu}_{\bm z}^{(t)}-\hat{\bm\mu}_{\bm z}^{(t+1)}\right)\\
&=n_t\hat{\bm\Sigma}_{\bm z}^{(t)}+n_t\left(\hat{\bm\mu}_{\bm z}^{(t)}-
\frac{n_t\hat{\bm\mu}_{\bm z}^{(t)}+n_b\hat{\bm\mu}_{\bm z}^{(b)}}{n_t+n_b}\right)^T\left(\hat{\bm\mu}_{\bm z}^{(t)}-\frac{n_t\hat{\bm\mu}_{\bm z}^{(t)}+n_b\hat{\bm\mu}_{\bm z}^{(b)}}{n_t+n_b}\right)\\
&=n_t\hat{\bm\Sigma}_{\bm z}^{(t)}+\frac{n_tn_b^2}{(n_t+n_b)^2}\left(\hat{\bm\mu}_{\bm z}^{(t)}
-\hat{\bm\mu}_{\bm z}^{(b)}\right)^T
\left(\hat{\bm\mu}_{\bm z}^{(t)}-\hat{\bm\mu}_{\bm z}^{(b)}\right)
\end{align*}
Applying the same techniques to the second sum term, we obtain
\begin{align*}
&\quad \sum_{i=n_t+1}^{n_t+n_b}\left(\bm{z}_i-\hat{\bm\mu}_{\bm z}^{(t+1)}\right)^T
\left(\bm{z}_i-\hat{\bm\mu}_{\bm z}^{(t+1)}\right)\\
&=\sum_{i=n_t+1}^{n_t+n_b}\left(\bm{z}_i-\hat{\bm\mu}_{\bm z}^{(b)}+\hat{\bm\mu}_{\bm z}^{(b)}-\hat{\bm\mu}_{\bm z}^{(t+1)}\right)^T\left(\bm{z}_i-\hat{\bm\mu}_{\bm z}^{(b)}+\hat{\bm\mu}_{\bm z}^{(b)}-\hat{\bm\mu}_{\bm z}^{(t+1)}\right)\\
&=\sum_{i=n_t+1}^{n_t+n_b}\left(\bm{z}_i-\hat{\bm\mu}_{\bm z}^{(b)}\right)^T
\left(\bm{z}_i-\hat{\bm\mu}_{\bm z}^{(b)}\right)+\sum_{i=n_t+1}^{n_t+n_b}\left(\hat{\bm\mu}_{\bm z}^{(b)}-\hat{\bm\mu}_{\bm z}^{(t+1)}\right)^T\left(\hat{\bm\mu}_{\bm z}^{(b)}-\hat{\bm\mu}_{\bm z}^{(t+1)}\right)\\
&\quad+\sum_{i=n_t+1}^{n_t+n_b}\left(\bm{z}_i-\hat{\bm\mu}_{\bm z}^{(b)}\right)^T
\left(\hat{\bm\mu}_{\bm z}^{(t)}-\hat{\bm\mu}_{\bm z}^{(t+1)}\right)+\sum_{i=n_t+1}^{n_t+n_b}\left(\hat{\bm\mu}_{\bm z}^{(b)}-\hat{\bm\mu}_{\bm z}^{(t+1)}\right)^T\left(\bm{z}_i-\hat{\bm\mu}_{\bm z}^{(t)}\right)\\
&=\sum_{i=n_t+1}^{n_t+n_b}\left(\bm{z}_i-\hat{\bm\mu}_{\bm z}^{(b)}\right)^T\left(\bm{z}_i-\hat{\bm\mu}_{\bm z}^{(b)}\right)+\sum_{i=n_t+1}^{n_t+n_b}\left(\hat{\bm\mu}_{\bm z}^{(b)}-\hat{\bm\mu}_{\bm z}^{(t+1)}\right)^T\left(\hat{\bm\mu}_{\bm z}^{(b)}-\hat{\bm\mu}_{\bm z}^{(t+1)}\right)\\
&=n_b\hat{\bm\Sigma}_{\bm z}^{(b)}+n_b\left(\hat{\bm\mu}_{\bm z}^{(b)}
-\hat{\bm\mu}_{\bm z}^{(t+1)}\right)^T\left(\hat{\bm\mu}_{\bm z}^{(b)}-\hat{\bm\mu}_{\bm z}^{(t+1)}\right)\\
&=n_b\hat{\bm\Sigma}_{\bm z}^{(b)}+n_b\left(\hat{\bm\mu}_{\bm z}^{(b)}
-\frac{n_t\hat{\bm\mu}_{\bm z}^{(t)}+n_b\hat{\bm\mu}_{\bm z}^{(b)}}{n_t+n_b}\right)^T\left(\hat{\bm\mu}_{\bm z}^{(b)}-\frac{n_t\hat{\bm\mu}_{\bm z}^{(t)}+n_b\hat{\bm\mu}_{\bm z}^{(b)}}{n_t+n_b}\right)\\
&=n_b\hat{\bm\Sigma}_{\bm z}^{(b)}+\frac{n_t^2n_b}{(n_t+n_b)^2}\left(\hat{\bm\mu}_{\bm z}^{(b)}-\hat{\bm\mu}_{\bm z}^{(t)}\right)^T\left(\hat{\bm\mu}_{\bm z}^{(b)}-\hat{\bm\mu}_{\bm z}^{(t)}\right)\\
\end{align*}
Therefore,
\begin{align}\label{update_cov}
\hat{\bm\Sigma}_{\bm z}^{(t+1)}&=\frac{1}{n_t+n_b}\left[n_t\hat{\bm\Sigma}_{\bm z}^{(t)}+\frac{n_tn_b^2}{(n_t+n_b)^2}\left(\hat{\bm\mu}_{\bm z}^{(t)}-\hat{\bm\mu}_{\bm z}^{(b)}\right)^T
\left(\hat{\bm\mu}_z^{(t)}-\hat{\bm\mu}_{\bm z}^{(b)}\right)+n_b\hat{\bm\Sigma}_{\bm z}^{(b)}\right.\nonumber\\
&\quad\left.+\frac{n_t^2n_b}{(n_t+n_b)^2}\left(\hat{\bm\mu}_{\bm z}^{(b)}
-\hat{\bm\mu}_{\bm z}^{(t)}\right)^T\left(\hat{\bm\mu}_{\bm z}^{(b)}-\hat{\bm\mu}_{\bm z}^{(t)}\right)\right]\nonumber\\
&=\frac{n_t}{n_t+n_b}\hat{\bm\Sigma}_{\bm z}^{(t)}+\frac{n_b}{n_t+n_b}\hat{\bm\Sigma}_{\bm z}^{b}
+\frac{n_tn_b}{(n_t+n_b)^2}\left(\hat{\bm\mu}_{\bm z}^{(t)}-\hat{\bm\mu}_{\bm z}^{(b)}\right)^T
\left(\hat{\bm\mu}_{\bm z}^{(t)}-\hat{\bm\mu}_{\bm z}^{(b)}\right)
\end{align}

\section{Mode}\label{mode_dev}
Note that the mean, variance, and mode of a log normal distribution with logarithm of location $\mu$ and logarithm of scale $\sigma$ are given by the following formulas:
\begin{align*}
\mu_z=e^{\mu+\frac{1}{2}\sigma^2}, \sigma_z^2=\left(e^{\sigma^2}-1\right)e^{2\mu+\sigma^2},
m_z = e^{\mu-\sigma^2}
\end{align*}
Then the mode $m_z$ can be expressed by the mean and variance,
\begin{align*}
m_z=e^{\mu+\frac{1}{2}\sigma^2-\frac{3}{2}\sigma^2}=\underbrace{e^{\mu+\frac{1}{2}\sigma^2}}_{\mu_z}e^{-\frac{3}{2}\sigma^2}
=\mu_z\bigg(\frac{\sigma_z^2}{\mu_z^2}+1\bigg)^{-\frac{3}{2}}
\end{align*}
as
\begin{align*}
&\frac{\sigma_{z}^{2}}{\mu_{z}^{2}}=
\frac{\left(e^{\sigma^{2}}-1\right)e^{2\mu+\sigma^{2}}}{\left(e^{\mu+\frac{1}{2}\sigma^{2}}\right)^{2}}
=\frac{\left(e^{\sigma^{2}}-1\right)e^{2\mu+\sigma^{2}}}{e^{2\mu+\sigma^{2}}}
=e^{\sigma^{2}}-1\\&\Rightarrow e^{\sigma^{2}}
=\frac{\sigma_{z}^{2}}{\mu_{z}^{2}}+1\Rightarrow e^{-\frac{3}{2}\sigma^{2}}
=\left(\frac{\sigma_{z}^{2}}{\mu_{z}^{2}}+1\right)^{-\frac{3}{2}}
\end{align*}

\section{Proof for Proposition 2}\label{proposition_proof}
Note that the marginal distribution of the reconstruction feature in $p_m$ is
$\frac{1}{2}p_0(z)+\frac{1}{2}p_t(z)$. For any $z\in[0,m_z]$, we have
\begin{align*}
\frac{1}{2}p_0(z)+\frac{1}{2}p_t(z)&=\frac{1}{2}p_0(z)+\frac{1}{2}\cdot\frac{\frac{1}{\sqrt{2\pi\sigma_z^2}}
e^{-\frac{(z-m_z)^2}{2\sigma_z^2}}}{\Phi(\frac{m_z-m_z}{\sigma_z})
-\Phi(\frac{0-m_z}{\sigma_z})}\\
&=\frac{1}{2}p_0(z)+\frac{1}{2}\cdot\frac{\frac{1}{\sqrt{2\pi\sigma_z^2}}
e^{-\frac{(z-m_z)^2}{2\sigma_z^2}}}{\frac{1}{2}
-\Phi(\frac{0-m_z}{\sigma_z})}\\
&\geq\frac{1}{2}p_0(z)+\frac{1}{2}\cdot\frac{\frac{1}{\sqrt{2\pi\sigma_z^2}}
e^{-\frac{(z-m_z)^2}{2\sigma_z^2}}}{\frac{1}{2}}\\
&=\frac{1}{2}p_0(z)+\frac{1}{\sqrt{2\pi\sigma_z^2}}
e^{-\frac{(z-m_z)^2}{2\sigma_z^2}}\\
&\geq\frac{1}{2}p_0(z)+\frac{1}{\sqrt{2\pi\sigma_z^2}}
e^{-\frac{(z-\mu_z)^2}{2\sigma_z^2}}\\
&\geq\frac{1}{\sqrt{2\pi\sigma_z^2}}
e^{-\frac{(z-\mu_z)^2}{2\sigma_z^2}}\\
&=p_n(z)
\end{align*}
where the second inequality is according to the fact $m_z\leq\mu_z$.

\section{Common Experimental Settings}
All experiments are conducted on an HP Z4 G4 Workstation equipped with an NVIDIA RTX 3090 GPU (24 GB VRAM). The software environment is configured using Conda with PyTorch 2.4.1 and Scikit-Learn 1.5.1.

For CANCE, the available training data are split into training and validation sets with an $80:20$ ratio. Both sets consist exclusively of normal samples, and the validation set is used for model selection. For NCE training, unless otherwise specified, the noise-to-sample ratio is fixed at $\nu=8$. All experiments are repeated five times, except for MNIST-C and KDDCUP99, for which we use 30 and 20 independent runs, respectively. Unless otherwise specified, performance is measured by AUROC and reported as the mean and standard deviation across runs.

Experiment-specific configurations for our method, as well as the settings and modifications used for individual baselines, are described separately in the corresponding experimental sections.

\section{Detailed Ablation Study}\label{ablation}
\paragraph{Composite feature via AE}

We use the same convolutional AE architecture for MNIST and F-MNIST and a higher-capacity convolutional AE for CIFAR-10, while keeping the architecture fixed across all compared variants within each dataset. The dataset-specific training configurations used throughout the ablation study are summarized in Table~\ref{tab:ablation-config}.

In the unimodal AD task, we construct each case from a labeled image dataset by treating one class as normal and all remaining classes as anomalies. Only normal samples from the training set are used for training and model selection, while the entire test set is used for evaluation. Table \ref{ablation-unimode} reports class-wise results on MNIST, F-MNIST, and CIFAR-10. LatNCE and CNCE have similar performance on CIFAR-10 dataset, however, CNCE significantly outperforms LatNCE across the $10$ classes under a one-sided Wilcoxon signed-rank test ($p=0.00488$). CANCE further outperforms CNCE in $17$ of the $20$ class-wise comparisons on MNIST and F-MNIST.

In multimodal AD task, two or more classes are treated as normal, with the same training and evaluation protocol.  Specifically, we consider three normal-class combinations, $(0,1)$, $(0,8)$, and $(1,8)$, with the remaining classes treated as anomalies. For MNIST-C, the entire MNIST training set is used for training, while at test time the original MNIST images are treated as normal and their corrupted counterparts in MNIST-C \citep{mu2019mnist} as anomalies. As shown in Table~\ref{ablation-multimodal}, CANCE achieves the best or near-best performance across these settings, demonstrating that the improvements observed in the unimodal setting extend to multimodal normal data.

\begin{table}[t]
\centering
\caption{Training configurations for the ablation experiments.}
\label{tab:ablation-config}
\begin{tabular}{lcccc}
\hline
Hyperparameter & MNIST & F-MNIST & CIFAR-10 & MNIST-C \\
\hline
\multicolumn{5}{l}{AE} \\
Input channels         & 1      & 1      & 3       & 1 \\
Hidden channels        & 8      & 8      & 32      & 8 \\
Latent dimension       & 6      & 6      & 64      & 10 \\
Activation             & GELU   & GELU   & GELU    & GELU \\
Learning rate          & $10^{-4}$ & $10^{-4}$ & $10^{-4}$ & $10^{-4}$ \\
Batch size             & 512    & 512    & 512     & 512 \\
$\lambda$              & 0.3  & 0.3    & 0.004   & 0.3 \\
Encoder epochs         & 400    & 200    & 20      & 400 \\
Decoder epochs         & 800    & 400    & 40      & 800 \\
\hline
\multicolumn{5}{l}{NCE estimator} \\
Noise learning rate    & $10^{-4}$ & $10^{-4}$ & $10^{-5}$ & $10^{-4}$ \\
Estimator learning rate & $10^{-4}$ & $10^{-4}$ & $5\times10^{-5}$ & $10^{-4}$ \\
Activation             & LeakyReLU & LeakyReLU & ELU & LeakyReLU \\
Batch size             & 64     & 64     & 64      & 64 \\
Hidden dimension       & 48     & 48     & 48      & 48 \\
Training epochs        & 250    & 250    & 50      & 250 \\
\hline
\end{tabular}
\end{table}

\begin{table}[!ht]
\caption{Ablation study on unimodal training dataset}
\label{ablation-unimode}
\centering
\begin{tabular}{cccccc}
\hline
Data &Error & LatNCE & CNCE & CANCE\\
\hline
\multicolumn{3}{l}{MNIST}\\
$0$ &$99.4(0.1)$ & $85.0(3.5)$ & $99.5(0.0)$ & $99.6(0.0)$ \\
$1$ &$99.9(0.0)$ & $98.4(0.3)$ & $99.8(0.0)$ & $99.8(0.0)$ \\
$2$ &$90.3(1.5)$ & $81.9(6.5)$ & $96.6(0.7)$ & $97.1(0.5)$ \\
$3$ &$92.6(0.6)$ & $81.2(1.2)$ & $95.6(0.4)$ & $96.7(0.3)$ \\
$4$ &$95.5(1.3)$ & $75.0(3.8)$ & $95.8(0.3)$ & $96.9(0.4)$ \\
$5$ &$95.1(1.5)$ & $76.1(4.8)$ & $96.1(0.4)$ & $97.2(0.4)$ \\
$6$ &$98.9(0.3)$ & $88.2(4.3)$ & $99.2(0.0)$ & $99.4(0.0)$ \\
$7$ &$96.1(0.6)$ & $89.1(2.1)$ & $96.9(0.6)$ & $97.5(0.3)$ \\
$8$ &$85.8(0.4)$ & $80.6(1.3)$ & $94.6(0.5)$ & $95.6(0.3)$ \\
$9$ &$97.0(0.3)$ & $86.0(1.5)$ & $96.2(0.2)$ & $97.1(0.1)$ \\
Avg &$95.1$ & $84.1$ & $97.0$ & $97.7$ \\
\hline
\multicolumn{3}{l}{F-MNIST}\\
$0$ &$89.9(0.1)$ & $88.9(0.5)$   & $90.0(0.7)$ & $90.6(0.4)$ \\
$1$ &$98.6(0.0)$ & $92.9(1.4)$ & $98.1(0.3)$ & $98.4(0.1)$ \\
$2$ &$87.1(0.3)$ & $88.2(0.6)$ & $89.1(0.6)$ & $89.0(0.4)$ \\
$3$ &$92.3(0.3)$ & $87.6(2.2)$ & $90.9(1.0)$ & $91.8(0.6)$ \\
$4$ &$90.0(0.3)$ & $86.9(0.8)$ & $89.8(0.3)$ & $90.7(0.2)$ \\
$5$ &$87.9(0.3)$ & $93.1(0.6)$ & $93.4(0.8)$ & $92.6(1.0)$ \\
$6$ &$82.1(0.4)$ & $80.6(2.1)$ & $82.7(0.4)$ & $83.5(0.2)$ \\
$7$ &$98.1(0.0)$ & $92.2(2.8)$ & $97.7(0.3)$ & $97.8(0.2)$ \\
$8$ &$82.3(0.3)$ & $90.7(0.5)$ & $91.9(0.9)$ & $91.7(1.1)$ \\
$9$ &$95.8(0.3)$ & $94.7(0.7)$ & $98.1(0.1)$ & $98.3(0.1)$ \\
Avg &$90.4$ &$89.6$  &$92.2$  &$92.5$ \\
\hline
\multicolumn{3}{l}{CIFAR-10}\\
$0$ &$57.8(2.3)$ & $63.4(2.0)$ & $63.4(2.3)$ & $63.8(2.3)$ \\
$1$ &$33.8(1.5)$ & $63.4(1.2)$ & $63.4(0.7)$ & $63.6(0.8)$ \\
$2$ &$65.0(0.4)$ & $59.5(1.8)$ & $59.9(1.8)$ & $60.0(1.7)$ \\
$3$ &$54.6(0.5)$ & $62.3(1.9)$ & $62.2(2.0)$ & $62.3(2.1)$ \\
$4$ &$71.0(0.8)$ & $68.9(1.6)$ & $69.3(1.7)$ & $69.4(1.5)$ \\
$5$ &$54.6(0.8)$ & $61.1(1.1)$ & $61.5(1.2)$ & $61.6(1.1)$ \\
$6$ &$55.2(2.9)$ & $73.0(2.4)$ & $73.0(2.4)$ & $73.0(2.3)$ \\
$7$ &$44.7(0.7)$ & $61.1(1.1)$ & $61.3(1.1)$ & $61.2(1.0)$ \\
$8$ &$67.8(0.8)$ & $71.9(2.1)$ & $72.4(2.1)$ & $72.3(2.3)$ \\
$9$ &$36.4(1.1)$ & $66.3(1.6)$ & $66.9(1.7)$ & $66.5(1.5)$ \\
Avg &$54.1$ & $65.1$ & $65.3$ & $65.4$ \\
\hline
\end{tabular}
\end{table}

\begin{table}[!ht]
\caption{Ablation study on the multimodal datasets}
\label{ablation-multimodal}
\centering
\begin{tabular}{ccccc}
\hline
 Data &Error &LatNCE &CNCE &CANCE\\
\hline
\multicolumn{5}{l}{MNIST}\\
$0,1$ &$99.4(0.1)$ & $93.8(1.1)$ & $99.5(0.1)$ & $99.6(0.1)$\\
$0,8$ &$87.8(0.7)$ & $79.8(1.3)$ & $95.0(0.5)$ & $95.5(0.4)$\\
$1,8$ &$96.7(0.7)$ & $89.3(2.3)$ & $97.4(0.3)$ & $97.8(0.2)$\\
Avg &$94.6$ & $87.6$ & $97.3$ & $97.6$\\
\hline
\multicolumn{5}{l}{F-MNIST}\\
$0,1$ &$93.0(0.2)$ &$91.7(0.8)$  &$92.6(0.5)$  &$92.7(0.4)$ \\
$0,8$ &$80.1(0.2)$ &$85.0(1.0)$ &$86.7(0.6)$  &$87.2(0.4)$ \\
$1,8$ &$86.9(0.4)$ &$90.2(2.5)$ &$92.5(0.8)$  &$92.5(0.6)$ \\
Avg &$86.7$ &$89.0$  &$90.6$  &$90.8$ \\
\hline
\multicolumn{5}{l}{CIFAR-10}\\
$0,1$ &$44.4(1.3)$ & $53.9(0.9)$ & $54.5(1.3)$ & $54.8(1.1)$\\
$0,8$ &$65.7(1.7)$ & $64.4(5.2)$ & $64.6(5.7)$ & $64.7(5.7)$\\
$1,8$ &$48.9(0.7)$ & $63.3(1.8)$ & $63.4(1.7)$ & $63.1(1.5)$\\
Avg &$53.0$ & $60.5$ & $60.8$ & $60.9$ \\
\hline
\multicolumn{5}{l}{Train: MNIST; Test: MNIST-C} \\
MNIST &$89.7(0.5)$ & $78.4(1.4)$ & $91.3(0.4)$ & $92.2(0.4)$  \\
\hline
\end{tabular}
\end{table}
\paragraph{Composite feature via ResNet followed by PCA}
For the ResNet-18 feature extraction shown in Table~\ref{features_ResNet-18_mnist_cifar10_full},
CNCE consistently outperforms LatNCE, highlighting the effectiveness of including reconstruction error in density estimation.
Additionally, augmenting indeed helps detect anomalies, as evidenced by the gap between CANCE and CNCE.
However, unlike in Table \ref{ablation-unimode}, LatNCE and CNCE are not competitive with Error on average.
There are two potential reasons for this: first, the latent feature extracted by PCA is simple and thus less useful compared to AE;
second, the density estimator induced by NCE does not capture data distribution well enough for AD.
However, CANCE outperforms Error, meaning there is still information in the latent representation.

\begin{table}[!ht]
\caption{Ablation study on features extracted by ResNet on MNIST, F-MNIST, and CIFAR-10}
\label{features_ResNet-18_mnist_cifar10_full}
\centering
\begin{tabular}{ccccc}
\hline
Data &Error  & LatNCE & CNCE & CANCE\\
\hline
\multicolumn{3}{l}{MNIST, ResNet-18\textbar{}ResNet-50}\\
0 & $98.1\mid96.3$ & $78.4(0.9)\mid75.2(0.4)$ & $96.7(0.6)\mid95.8(0.4)$ & $98.6(0.0)\mid97.4(0.1)$ \\
1 & $99.8\mid98.6$ & $98.6(0.1)\mid95.9(0.6)$ & $99.6(0.0)\mid98.5(0.1)$ & $99.7(0.0)\mid98.5(0.1)$ \\
2 & $88.3\mid76.6$ & $74.4(0.7)\mid66.1(0.8)$ & $83.8(1.4)\mid80.2(1.0)$ & $91.0(0.1)\mid83.9(0.2)$ \\
3 & $95.1\mid95.9$ & $74.1(0.5)\mid90.7(0.1)$ & $92.5(0.3)\mid94.8(0.1)$ & $94.7(0.2)\mid95.5(0.1)$ \\
4 & $97.3\mid95.2$ & $84.4(0.5)\mid81.7(0.8)$ & $96.3(0.3)\mid94.1(0.1)$ & $97.3(0.2)\mid95.2(0.2)$ \\
5 & $92.2\mid87.3$ & $60.8(0.2)\mid72.5(0.4)$ & $87.7(0.5)\mid85.3(0.2)$ & $91.5(0.4)\mid87.8(0.1)$ \\
6 & $94.8\mid88.1$ & $72.7(0.8)\mid83.5(0.5)$ & $93.5(0.2)\mid89.9(0.5)$ & $94.6(0.2)\mid91.9(0.3)$ \\
7 & $96.8\mid90.3$ & $85.5(1.2)\mid68.7(0.9)$ & $96.3(0.3)\mid86.9(0.4)$ & $96.8(0.3)\mid89.9(0.2)$ \\
8 & $91.0\mid89.3$ & $75.4(1.1)\mid86.4(0.5)$ & $85.8(1.5)\mid91.1(0.3)$ & $92.1(0.5)\mid92.5(0.3)$ \\
9 & $92.0\mid89.4$ & $65.5(0.6)\mid73.5(0.1)$ & $89.1(0.4)\mid86.5(1.9)$ & $93.3(0.2)\mid90.0(1.1)$ \\
Avg & $94.5\mid90.7$ & $77.0\mid79.4$ & $92\mid90.3$ & $95.0\mid92.2$ \\
\hline
\multicolumn{3}{l}{F-MNIST, ResNet-18\textbar{}ResNet-50}\\
0 & $88.7\mid89.9$ & $73.8(1.2)\mid72.3(1.2)$ & $85.6(0.9)\mid89.6(0.4)$ & $89.7(0.3)\mid91.3(0.3)$ \\
1 & $98.2\mid98.3$ & $95.3(0.2)\mid84.1(0.7)$ & $98.5(0.0)\mid98.5(0.1)$ & $98.5(0.1)\mid98.7(0.0)$ \\
2 & $90.2\mid91.5$ & $73.3(1.0)\mid63.3(1.3)$ & $89.7(0.1)\mid91.6(0.2)$ & $91.0(0.2)\mid92.6(0.1)$ \\
3 & $90.5\mid91.5$ & $62.1(2.0)\mid66.9(0.8)$ & $82.3(0.9)\mid86.2(1.9)$ & $85.7(0.2)\mid90.1(0.5)$ \\
4 & $90.9\mid87.7$ & $65.6(0.8)\mid80.6(0.7)$ & $90.7(0.1)\mid89.4(0.5)$ & $92.3(0.2)\mid91.7(0.1)$ \\
5 & $97.4\mid96.3$ & $74.8(1.2)\mid73.2(0.6)$ & $95.9(0.4)\mid95.1(0.3)$ & $96.8(0.2)\mid95.8(0.2)$ \\
6 & $81.0\mid78.6$ & $64.6(2.1)\mid66.6(1.9)$ & $79.4(0.3)\mid79.1(1.1)$ & $81.1(0.6)\mid82.6(0.2)$ \\
7 & $96.8\mid97.8$ & $91.9(0.5)\mid88.2(0.2)$ & $97.0(0.1)\mid97.7(0.1)$ & $97.4(0.1)\mid98.0(0.1)$ \\
8 & $89.2\mid91.2$ & $82.0(2.1)\mid74.2(0.8)$ & $92.6(0.2)\mid91.6(1.5)$ & $94.9(0.6)\mid94.0(0.2)$ \\
9 & $97.5\mid97.5$ & $83.6(0.7)\mid80.7(0.3)$ & $97.8(0.2)\mid96.5(0.2)$ & $98.2(0.1)\mid97.5(0.1)$ \\
Avg & $92.1\mid92.1$ & $76.7\mid75.0$ & $90.9\mid91.5$ & $92.6\mid93.2$ \\
\hline
\multicolumn{3}{l}{CIFAR-10, ResNet-18\textbar{}ResNet-50}\\
0 & $88.5\mid92.3$ & $75.5(0.8)\mid82.8(1.1)$ & $80.9(1.6)\mid88.2(1.2)$ & $86.5(0.4)\mid90.4(0.2)$ \\
1 & $94.6\mid95.8$ & $90.9(0.3)\mid93.8(0.2)$ & $93.5(0.6)\mid95.4(0.2)$ & $95.3(0.1)\mid95.9(0.1)$ \\
2 & $77.0\mid83.5$ & $60.0(0.2)\mid64.9(0.4)$ & $63.9(3.1)\mid72.9(1.6)$ & $75.1(0.3)\mid81.2(0.4)$ \\
3 & $80.3\mid86.1$ & $71.3(1.6)\mid76.8(0.6)$ & $75.1(1.7)\mid80.6(1.3)$ & $78.8(0.7)\mid84.9(0.3)$ \\
4 & $90.2\mid93.1$ & $81.2(0.8)\mid85.2(0.4)$ & $85.8(1.0)\mid89.6(1.5)$ & $89.2(0.3)\mid91.8(0.2)$ \\
5 & $84.1\mid88.7$ & $68.6(1.2)\mid78.9(1.3)$ & $76.9(2.7)\mid86.3(2.0)$ & $87.0(0.4)\mid90.3(0.2)$ \\
6 & $90.4\mid93.2$ & $80.5(0.7)\mid84.3(0.2)$ & $85.7(0.9)\mid88.5(0.7)$ & $88.9(0.9)\mid92.1(0.3)$ \\
7 & $86.5\mid90.3$ & $76.0(0.5)\mid79.5(0.6)$ & $85.4(2.7)\mid87.2(2.6)$ & $91.1(0.4)\mid93.3(0.3)$ \\
8 & $91.7\mid94.9$ & $83.1(0.6)\mid89.2(0.4)$ & $88.7(0.3)\mid93.0(0.7)$ & $92.4(0.3)\mid94.7(0.1)$ \\
9 & $94.6\mid95.8$ & $88.7(0.9)\mid92.1(0.3)$ & $93.4(0.8)\mid94.9(0.4)$ & $95.8(0.1)\mid96.2(0.1)$ \\
Avg & $87.8\mid91.4$ & $77.6\mid82.7$ & $82.9\mid87.7$ & $88.0\mid91.1$ \\
\hline
\end{tabular}
\end{table}

To examine the impact of pretrained models on CANCE’s performance, we use ResNet-50 for feature extraction with PCA on MNIST, F-MNIST, and CIFAR-10, keeping the same architecture and hyper-parameters as with ResNet-18. The corresponding ablation results are included in Table \ref{features_ResNet-18_mnist_cifar10_full}.

\section{Detailed Benchmark Comparison on ADBench}\label{sec:adbench_detailed}
We evaluate CANCE on the ADBench benchmark \citep{han2022adbench}, which comprises a diverse collection of datasets spanning tabular, image, and natural language data.\footnote{In ADBench, image and natural language datasets are represented by features extracted using pretrained ResNet-18 and BERT models, respectively.} For direct comparison, we use the same datasets and dataset preparation as DTE \citep{livernoche2023diffusion}. For CANCE, each feature of the tabular and natural language datasets is standardized to have zero mean and unit variance, while no additional preprocessing or normalization is applied to the image datasets. We compare CANCE with DTE and several other baseline methods. All baseline results are taken directly from Table~13 of \citet{livernoche2023diffusion}, whereas the results for CANCE are obtained from our own experiments.

\paragraph{Implementation Details.}
In the first stage, the AE is trained for $800$ epochs using the AdamW optimizer with a learning rate of $10^{-4}$ and a batch size of $128$, and a regularization coefficient of $\lambda=0.1$. Both the encoder and decoder are updated during the first $400$ epochs, after which the encoder is fixed and only the decoder is updated for the remaining $400$ epochs. The AE adopts a symmetric architecture. Given an input dimension $d_0$, the latent dimension is set to $d_l=\min(64,\max(2,\lceil d_0/2\rceil))$, and the hidden-layer dimension is set to $d_h=\min(128,2d_l)$. Accordingly, the encoder has dimensions $d_0\rightarrow d_h\rightarrow d_h\rightarrow d_l$, with the decoder following the symmetric reverse structure. GELU activation is applied after each hidden layer.

In the second stage, the trained AE is fixed and used to construct the composite features. NCE is then applied to the composite features for density estimation, with an estimator network modeling the data density and an affine network optimizing the Gaussian noise distribution. The estimator and affine networks are trained for $3000$
using the AdamW optimizer with an initial learning rate of $10^{-4}$ and a batch size of $64$. The learning rates are scheduled using cosine annealing with $T_{\max}=3000$.

For the estimator network, the input dimension is $d_c=d_l+2$, corresponding to the latent representation augmented with the two reconstruction features. Its hidden-layer width is set to$d_h=\min(128,\max(8,2^{\lceil\log_2 d_c\rceil}))$.
The estimator network therefore has dimensions $d_c\rightarrow d_h\rightarrow d_h/2\rightarrow 1$, with a SiLU activation applied after each hidden layer.

\begin{table*}[!ht]
\caption{Average AUROC and standard deviations on ADBench dataset.}
\label{adbench}
\centering
{\scriptsize
\begin{tabular}{lccccccccc}
\hline
 & DAGMM & DSVDD & DROCC & GOAD & DDPM & DTE-NP & DTE-IG & DTE-C  & CANCE  \\
\hline
aloi &50.84(2.97)  &50.89(2.05)  &50.0(0.0)  &48.01(0.92)  &49.91(0.35)  &51.19(0.46) &50.87(1.17) &50.44(0.19) &52.33(0.97)\\
amazon  &50.47(2.01)  &51.2(4.42)  &50.0(0.0)  &56.07(0.9)  &55.09(0.1)  &60.82(0.0) &51.93(3.5)  &56.71(2.15)  &61.49(0.79)\\
\makecell[l]{annthy-\\roid} &72.23(15.13)  &55.01(3.62) &88.9(2.31) &81.01(5.16)  &88.82(1.34)  &92.9(0.25)  &87.6(4.46)  &97.52(0.15)  &88.19(1.58) \\
backdoor &54.36(19.91) &91.14(2.62) &94.25(0.73) &52.9(14.48)  &80.93(0.56) &99.31(1.7) &94.02(1.46) &91.65(1.73) &91.85(0.63) \\
breastw &89.53(10.2) &96.96(0.92) &47.32(31.95) &98.86(0.33) &98.7(0.43) &99.28(0.12) &78.65(11.1) &92.78(1.75) &99.52(0.09) \\
campaign &61.47(2.73) &62.21(12.88) &50.0(0.0) &47.89(12.32)  &74.51(0.42) &78.79(0.25)  &74.81(1.69)  &68.84(6.50) &68.84(6.50) \\
cardio &77.92(8.85) &65.43(4.37) &62.14(23.76) &96.01(0.27) &86.94(1.96)  &91.8(0.59)  &73.79(14.09)  &94.91(0.63) &94.91(0.63) \\
\makecell[l]{cardioto-\\cography} &67.11(9.03) &47.75(8.59) &45.98(16.75) &76.06(1.4) &54.45(2.96)  &63.76(1.88)  &52.44(3.67)  &60.13(2.59)  &76.96(1.31) \\
celeba &63.81(4.28)  &56.17(22.46) &68.89(1.11) &43.8(10.49)  &78.56(1.99)  &70.40(0.37)  &74.51(2.39)  &82.18(2.38)  &68.49(4.12) \\
census &52.24(1.19) &54.16(4.33) &55.36(3.62) &35.24(4.19) &70.15(0.23)  &72.1(0.4)  &61.79(4.94)  &69.62(0.91)  &66.88(2.64) \\
cover &75.94(14.06) &49.12(14.74) &95.79(0.69) &13.83(13.34) &98.35(0.66)  &97.73(0.55)  &95.83(1.55)  &97.76(1.28)  & 96.97(0.96) \\
donor &62.15(16.19) &72.95(17.81) &74.18(22.09) &33.57(16.0) &82.5(1.83)  &99.26(0.29)  &99.25(0.6)  &98.15(0.37)  &91.47(2.00) \\
fault &52.85(7.19) &54.31(1.62) &55.73(5.33) &58.89(0.61) &61.09(1.16)  &58.64(0.65)  &59.42(1.46)  &59.46(1.45)  &66.11(3.79) \\
fraud &85.33(6.76) &83.13(6.6) &50.0(0.0) &69.75(21.3) &93.65(0.92)  &95.64(1.05)  &90.79(3.24)  &93.52(1.53)  &99.41(0.08) \\
glass &65.33(15.53) &83.67(16.2) &64.89(23.64) &59.03(12.64) &66.67(13.63)  &89.64(3.54)  &98.53(0.86)  &92.42(2.3)  &79.60(3.70) \\
hepatitis &70.22(6.66) &99.57(0.24) &51.8(17.93) &84.5(3.25) &97.74(1.18) &93.22(3.9)  &99.93(0.15)  &98.78(0.88)  &93.18(2.70) \\
http &91.78(17.91) &61.31(51.49) &50.0(0.0) &99.68(0.13) &100.0(0.0)  &99.98(0.03)  &80.72(43.08)  &99.45(0.1) &99.86(0.09) \\
imdb &48.6(0.38)  &49.97(5.71)  &51.35(2.05)  &48.46(0.65)   &47.91(0.1)  &50.43(0.0) &50.97(3.41) &48.05(2.22)  &51.62(0.66) \\
\makecell[l]{internet-\\ads} &49.47(5.05) &72.96(3.24) &53.4(7.55) &65.65(0.21) &65.76(0.06)  &69.96(2.22) &71.52(3.89) &77.57(1.54) &87.28(0.17) \\
ionosphere &73.95(5.98) &97.2(1.26) &61.14(28.54) &91.54(3.05) &94.6(0.85)  &97.77(1.39) &95.15(3.6) &95.42(0.58) &97.36(0.26) \\
landsat &56.27(3.46) &59.44(1.41) &53.86(2.57) &40.52(2.32) &51.37(1.0) &68.2(1.75) &44.72(5.52)  &52.79(1.63) &65.79(2.19) \\
letter &38.97(8.38) &36.4(3.05) &55.26(11.04) &31.08(0.55) &38.05(1.12) &34.38(0.98) &39.86(2.29) &36.72(0.95) &83.53(2.80) \\
\makecell[l]{lympho-\\graph} &94.94(3.85) &99.73(0.3) &32.42(37.75) &99.89(0.08) &99.94(0.09) &99.93(0.09) &99.98(0.05) &98.99(0.4) &99.79(0.08) \\
\makecell[l]{magic.-\\gamma} &59.23(4.32) &62.97(1.07) &78.83(0.66) &69.46(2.39) &85.97(1.08) &83.57(0.76) &86.46(1.12)  &87.5(0.9) &76.68(1.47) \\
\makecell[l]{mammo-\\graphy} &76.03(14.65) &71.5(7.4) &81.82(1.94) &69.94(8.59) &81.01(2.04) &87.62(0.09) &84.64(3.47) &86.42(1.72) &84.98(0.80) \\
mnist &72.19(7.16) &66.37(11.03) &83.13(1.64) &90.07(0.35) &87.27(3.22) &94.02(0.42) &80.78(5.91) &87.43(2.48) &91.41(1.51) \\
musk &95.01(4.27) &99.99(0.01) &32.99(33.18) &100.0(0.0) &100.0(0.0) &100.0(0.0) &94.22(12.93) &100(0.0) &100.00(0.00) \\
optdigits &40.04(20.45) &39.45(18.65) &85.25(2.9) &67.46(4.94) &90.76(2.12) &94.28(1.67) &79.81(9.35) &82.38(2.68) &73.68(6.21) \\
pageblocks &82.8(10.44) &78.39(1.53) &92.33(1.18) &88.05(1.19) &86.93(0.42) &89.32(0.26) &85.66(1.77) &89.89(0.55)  &96.02(0.16) \\
pendigits &56.49(21.83) &46.29(11.35) &75.91(13.38) &89.97(2.03) &98.11(0.23) &99.61(0.17) &96.96(1.63) &97.79(0.68) &98.86(0.61) \\
pima &54.54(6.0) &57.99(2.85) &47.53(16.52) &62.31(13.7) &70.27(2.28) &81.5(2.57) &68.59(3.93) &69.88(1.95) &68.79(1.39) \\
satellite &72.79(2.01) &76.19(2.66) &73.38(4.44) &68.76(0.92) &77.7(0.53) &82.11(0.67) &76.52(2.68) &78.61(0.7)  &83.61(1.44) \\
satimage-2 &91.82(4.43) &92.94(2.84) &99.22(0.42) &98.99(0.06) &99.62(0.16) &99.67(0.0) &95.34(1.84) &99.34(0.07) &99.52(0.06) \\
shuttle &84.58(18.68) &99.79(0.07) &50.0(0.0) &70.44(16.28) &99.91(0.01) &99.93(0.02) &99.86(0.12) &99.75(0.0) &99.76(0.03) \\
skin &67.91(30.02) &59.95(4.47) &89.48(1.07) &64.95(2.25) &88.74(4.4) &98.86(0.46) &98.71(1.13) &91.77(0.22) &90.81(0.75) \\
smtp &87.08(5.31) &85.24(6.6) &57.13(15.93) &78.78(13.12) &95.43(1.26) &92.98(2.91) &81.64(9.87) &95.27(1.28) &97.85(0.66) \\
spambase &69.41(4.42) &70.24(5.02) &75.37(4.45) &81.78(0.36) &64.54(0.8) &83.74(0.69) &77.5(3.37) &83.01(0.41) &80.55(0.99) \\
speech &50.66(3.9) &48.88(2.88) &48.96(2.21) &36.63(1.14) &36.96(0.86) &41.37(0.0) &39.57(1.54) &38.17(0.57) &43.28(4.44) \\
stamps &80.11(11.56) &71.09(3.68) &50.15(22.93) &81.46(15.36) &91.84(4.15) &97.87(0.37) &93.38(3.31) &91.6(2.04) &91.53(2.35) \\
thyroid &91.08(6.68) &88.77(3.69) &94.96(1.55) &95.15(0.85) &97.95(0.22) &98.63(0.04) &89.43(11.73) &98.74(0.14) &98.06(0.31) \\
vertebral &50.6(11.93) &44.79(1.71) &43.84(24.28) &46.73(8.74) &70.67(6.28) &54.3(15.47) &74.63(7.58) &66.41(1.9) &45.52(12.12) \\
vowels &42.55(11.56) &55.73(4.43) &54.74(21.32) &68.49(3.71) &86.38(1.94) &81.42(1.44) &85.68(3.16) &86.93(2.25) &87.06(5.41) \\
waveform &51.89(7.39) &59.94(2.58) &67.7(5.0) &64.99(3.1) &62.17(2.64) &74.48(0.0) &73.68(2.68) &65.21(1.05) &81.20(0.81) \\
wbc &86.76(15.33) &91.44(3.64) &44.24(28.21) &99.14(0.2) &99.2(0.39) &99.54(0.28) &90.98(11.64) &80.53(7.15) &98.91(0.34) \\
wdbc &73.78(25.27) &99.31(0.4) &40.08(33.97) &98.96(0.25) &99.3(0.17) &99.52(0.38) &99.6(0.31) &98.48(0.77)  &99.23(0.31) \\
wilt &41.81(7.29) &34.41(1.7) &49.49(12.62) &51.38(5.33) &71.66(0.81) &62.91(5.59) &93.75(3.23) &85.1(1.13) &47.15(7.98) \\
wine &66.17(39.61) &92.16(4.34) &43.8(32.33) &94.11(1.86) &99.61(0.09) &99.44(0.97) &99.95(0.11) &99.97(0.04) &97.69(2.21) \\
wpbc &46.99(2.56) &82.67(5.47) &43.78(4.34) &51.39(5.74) &66.49(3.17) &83.16(13.46) &70.71(9.06) &68.88(2.7) &62.79(0.90) \\
yeast &51.03(3.92) &47.62(5.96) &48.42(5.46) &52.53(3.88) &49.19(2.85) &44.58(0.33) &48.58(2.89) &47.08(1.1) &47.56(3.63) \\
yelp &49.87(1.25) &49.9(3.3) &50.67(1.15) &61.07(0.98) &59.3(0.09) &68.66(0.0) &57.32(4.3) &59.94(5.32) &68.85(0.76) \\
MNIST-C &63.71(6.37) &64.7(4.6) &57.2(11.1) &79.34(0.41) &80.12(0.14) &84.74(0.0) &79.92(3.93) &86.1(0.84) &86.37(0.60) \\
FashionMNIS &70.8(4.89) &75.45(2.22) &51.58(14.44) &88.03(0.24) &88.52(0.07) &90.14(0.0) &84.32(2.44) &90.21(0.55) &91.96(0.26) \\
CIFAR10 &53.97(2.96) &56.12(2.2) &49.63(3.46) &67.53(0.72) &67.91(0.13) &67.82(0.0) &62.4(3.33) &68.53(1.59) &70.24(0.33) \\
SVHN &53.36(2.12) &53.92(3.05) &49.99(2.47) &60.75(0.49) &61.37(0.008) &62.13(0.0) &59.19(2.5) &62.91(1.1) &63.77(0.25) \\
MVTec-AD &64.69(5.63) &89.55(2.19) &60.52(11.42) &77.06(2.1) &78.0(1.99) &89.66(1.51) &85.88(3.63) &89.39(2.0) &91.32(0.11) \\
20news &51.48(4.08) &55.64(4.17) &52.77(4.09) &54.92(1.04) & 54.88(0.52) &59.97(0.74) &58.28(6.15) &64.3(3.17) &64.15(0.63) \\
agnews &51.03(3.6) &49.83(5.63) &49.65(0.76) &59.87(0.86) &57.82(0.12) &67.98(0.0) &56.55(4.15) &68.18(3.22) &69.62(2.09) \\
\hline
\end{tabular}
}\end{table*}

\section{Additional Comparisons}\label{more_results}
This section provides additional comparisons of CANCE with three baseline methods, LSA, DAGMM, and DPAD.

\paragraph{LSA and DAGMM}
Given the connection between LSA and joint density estimation on the composite feature discussed in the main paper, we include LSA as a baseline for direct comparison. We additionally compare with DAGMM, which similarly combines latent representations and reconstruction features for density-based anomaly detection. We use the same implementation and hyperparameter settings as in the ablation study. The class-wise comparison results on MNIST and CIFAR-10 are reported in Table~\ref{lsa_comparision}. The results for DAGMM and CANCE are obtained from our own experiments, while those for all other methods are taken from Table 1 of LSA paper \citep{abati2019latent}. DAGMM is not evaluated on MNIST and CIFAR-10, we use the same architecture as our method, and the number of Gaussians within the model is set to $4$. However, on CIFAR-10, we have encountered issues with degenerated covariance matrices in the DAGMM. Therefore, we have changed the latent dimension from $64$ to $16$. In the process of implementing DAGMM, we find that DAGMM does not train stably if the latent dimension or the number of Gaussians within the model is not properly chosen.

Most methods, except DAGMM and Pix-CNN, perform favorably on MNIST. DAGMM performs poorly in this setting, consistent with previously reported difficulties when applying it to image datasets \citep{hojjati2023dasvdd}. Pix-CNN also yields relatively poor performance, which is consistent with our earlier discussion that the effective dimensionality of image data can be substantially lower than the ambient image dimensionality, making direct density estimation in the original image space challenging. Notably, the deep probabilistic models, including VAE, LSA, and CANCE, outperform KDE on MNIST. CIFAR-10 presents a substantially more challenging setting, likely due to its greater visual diversity and more complex backgrounds. Nevertheless, CANCE achieves higher average AUROC than LSA on both MNIST and CIFAR-10, while also outperforming LSA on the majority of individual classes.

\begin{table}[!ht]
\caption{AUCROC [\%] for baseline methods, Pix-CNN (PC) and DAGMM, compared to CANCE mean and std. value across $5$ independent runs. }
\label{lsa_comparision}
\centering
\begin{tabular}{cccccc|cc}
\hline
Data &KDE &VAE &PC  &LSA  &DAGMM  &CANCE\\
\hline
\multicolumn{3}{l}{MNIST}\\
$0$ &$88.5$  &$99.8$ & $53.1$ &$99.3$ & $53.6$ & $99.6(0.0)$\\
$1$ &$99.6$  &$99.9$ & $99.5$ &$99.9$ & $51.5$ & $99.8(0.0)$\\
$2$ &$71.0$  &$96.2$ & $47.6$ &$95.9$ & $53.5$ & $97.1(0.5)$\\
$3$ &$69.3$  &$94.7$ & $51.7$ &$96.6$ & $49.7$ & $96.7(0.3)$\\
$4$ &$84.4$  &$96.5$ & $73.9$ &$95.6$ & $52.7$ & $96.9(0.4)$\\
$5$ &$77.6$  &$96.3$ & $54.2$ &$96.4$ & $54.3$ & $97.2(0.4)$\\
$6$ &$86.1$  &$99.5$ & $59.2$ &$99.4$ & $55.2$ & $99.4(0.0)$\\
$7$ &$88.4$  &$97.4$ & $78.9$ &$98.0$ & $53.8$ & $97.5(0.3)$\\
$8$ &$66.9$  &$90.5$ & $34.0$ &$95.3$ & $54.8$ & $95.6(0.3)$\\
$9$ &$82.5$  &$97.8$ & $66.2$ &$98.1$ & $51.8$ & $97.1(0.1)$\\
Avg &$81.4$  &$96.9$ & $61.8$ &$97.5$ & $53.1$ & $97.7$\\
\hline
\multicolumn{3}{l}{CIFAR-10}\\
$0$ &$65.8$ &$68.8$ & $78.8$ &$73.5$ & $47.5$ & $63.8(2.3)$\\
$1$ &$52.0$ &$40.3$ & $42.8$ &$58.0$ & $47.2$ & $63.6(0.8)$\\
$2$ &$65.7$ &$67.9$ & $61.7$ &$69.0$ & $46.1$ & $60.0(1.7)$\\
$3$ &$49.7$ &$52.8$ & $57.4$ &$54.2$ & $47.3$ & $62.3(2.1)$\\
$4$ &$72.7$ &$74.8$ & $51.1$ &$76.1$ & $48.5$ & $69.4(1.5)$\\
$5$ &$49.6$ &$51.9$ & $57.1$ &$54.6$ & $48.9$ & $61.6(1.1)$\\
$6$ &$75.8$ &$69.5$ & $42.2$ &$75.1$ & $47.8$ & $73.0(2.3)$\\
$7$ &$56.4$ &$50.0$ & $45.4$ &$53.5$ & $47.6$ & $61.2(1.0)$\\
$8$ &$68.0$ &$70.0$ & $71.5$ &$71.7$ & $48.4$ & $72.3(2.3)$\\
$9$ &$54.0$ &$39.8$ & $42.6$ &$54.8$ & $48.1$ & $66.5(1.5)$\\
Avg &$61.0$ &$58.6$ & $55.1$ &$64.1$ & $47.7$ & $65.4$\\
\hline
\end{tabular}
\end{table}

We further conduct a dedicated comparison with DAGMM on KDDCUP99, a dataset used in the original DAGMM study \citep{zong2018deep}, to provide a direct comparison between the two methods. In this experiment, we use the same neural networks and training strategy as in DAGMM except three modifications: 1) DAGMM is only trained over $10$ epochs instead of $200$ epochs; 2) the reconstruction error is $\frac{\|\bm x-\bm x'\|^2}{d_0}$ rather than the relative Euclidean distance $\frac{\|\bm x-\bm x'\|}{\|\bm x\|}$; and the cosine dissimilarity $\frac{1}{2}\left(1-\frac{\bm x^T\bm x'}{\|\bm x\|\|\bm x'\|}\right)$ is utilized rather than the cosine similarity $\frac{\bm x^T\bm x'}{\|\bm x\|\|\bm x'\|}$. We observe that the training loss of DAGMM converges in less than $10$ epochs, making it unnecessary to train the model for $200$ epochs. Since the reconstruction error is used as an anomaly score instead of the relative Euclidean distance in AE, it is more reasonable to include the reconstruction error as a component of the composite feature. The reconstruction error is scaled by the data dimension $d_0$ to ensure its value remains small. Additionally, we choose cosine dissimilarity over cosine similarity because it is non-negative, similar to the reconstruction error. We independently run the experiment $20$ times \footnote{Refer to the DAGMM paper for the implementation details.}, as done in DAGMM, and summarize the average precision, recall, and $F_1$ score in Table \ref{dagmm-nce}. The values for DAGMM-0 are extracted from Table 2 in the DAGMM paper. The results for DAGMM-1 were obtained by training DAGMM with three modifications as mentioned earlier. The last row, DAGMM-CANCE, involves training a DAGMM first for feature learning and then using our method, CANCE, for density estimation. Clearly, DAGMM-1 achieves better performance than DAGMM-0. Furthermore, DAGMM-CANCE achieves comparable results to DAGMM-1. We argue that separately optimizing the compression network and estimation network will not degrade the method’s performance, provided they are well designed and optimized.

\begin{table}[!ht]
\caption{AD results on contaminated training data from KDDCUP99}
\label{dagmm-nce}
\begin{center}
\begin{tabular}{lccc}
\hline
Method & Precision & Recall & $F_1$\\
\hline
DAGMM-0 & $93.0$ & $94.4$ & $93.7$  \\
DAGMM-1 & $97.7(0.3)$  & $96.9(0.6)$  & $97.3(0.5)$ \\
DAGMM-CANCE & $97.6(0.3)$  & $96.9(0.6)$ & $97.3(0.5)$ \\
\hline
\end{tabular}
\end{center}
\end{table}

\paragraph{DPAD}
DPAD \citep{fu2024dense} is a representative one-class classification method for unsupervised AD with minimal data preprocessing. We therefore include DPAD as a baseline for comparison with CANCE. The comparison is conducted on Fashion-MNIST and CIFAR-10. For CANCE, we use the same implementation and hyperparameter settings as in the ablation study. In addition to the original AE-based configuration, we also consider pretrained ResNet-18 and ResNet-50 for feature extraction, followed by PCA for dimensionality reduction. The PCA dimensions are set to 6 for Fashion-MNIST and 64 for CIFAR-10, matching the latent dimensions of the corresponding AEs. The resulting PCA representations and reconstruction errors are used to construct the composite features, followed by the same NCE training procedure as in the original CANCE. The comparison results are reported in Table~\ref{fmnist_dpad}, where CANCE, CANCE18, and CANCE50 denote the AE-, ResNet-18-with-PCA-, and ResNet-50-with-PCA-based configurations of CANCE, respectively. Except for these three CANCE configurations, all results are taken from Tables 2 and 3 of the DPAD paper \citep{fu2024dense}.

As shown in Table~\ref{fmnist_dpad}, the original CANCE achieves slightly lower average AUROC than DPAD on Fashion-MNIST, while outperforming the other baselines. Incorporating pretrained representations further improves its performance. In particular, CANCE50 achieves an average AUROC of 93.2 on Fashion-MNIST, approaching the 93.7 achieved by DPAD. The improvement is particularly substantial on CIFAR-10, where the average AUROC increases from 65.4 with the AE-based CANCE to 88.0 and 91.1 with CANCE18 and CANCE50, respectively, both substantially outperforming DPAD at 74.5. These results show that CANCE can effectively leverage pretrained representations.

\begin{table}[t]
\caption{AUROC[\%] CANCE vs DPAD on F-MNIST and CIFAR-10. NN (nearest neighbor, use raw \textbar{} ResNet-18 feature), CANCE (AE), CANCE18 (ResNet-18), CANCE50 (ResNet-50)}
\label{fmnist_dpad}
\centering
\begin{tabular}{cccccccccc}
\hline
Data & DAGMM & DSVDD & NN  & DROCC  & DPAD  & CANCE & CANCE18 & CANCE50\\
\hline
\multicolumn{3}{l}{F-MNIST}\\
0 & $42.1$ & $79.1$ & $91.2\mid100$ & $88.1$ & $93.7(0.2)$ &  $90.6(0.4)$ & $89.7(0.3)$ & $91.3(0.3)$\\
1 & $55.1$ & $94.0$ & $98.6\mid100$ & $97.7$ & $98.7(0.0)$ &  $98.4(0.1)$ & $98.5(0.1)$ & $98.7(0.0)$\\
2 & $50.4$ & $83.0$ & $88.5\mid100$ & $87.6$ & $90.3(0.0)$ &  $89.0(0.4)$ & $91.0(0.2)$ & $92.6(0.1)$\\
3 & $57.0$ & $82.9$ & $93.6\mid100$ & $87.7$ & $94.7(0.3)$ &  $91.8(0.6)$ & $85.7(0.2)$ & $90.1(0.5)$\\
4 & $26.9$ & $87.0$ & $89.4\mid100$ & $87.2$ & $92.2(0.1)$ &  $90.7(0.2)$ & $92.3(0.2)$ & $91.7(0.1)$\\
5 & $70.5$ & $80.3$ & $90.0\mid100$ & $91.0$ & $93.9(0.8)$ &  $92.6(1.0)$ & $96.8(0.2)$ & $95.8(0.2)$\\
6 & $48.3$ & $74.9$ & $81.7\mid100$ & $77.1$ & $82.3(0.1)$ &  $83.5(0.2)$ & $81.1(0.6)$ & $82.6(0.2)$\\
7 & $83.5$ & $94.2$ & $98.4\mid100$ & $95.3$ & $98.7(0.1)$ &  $97.8(0.2)$ & $97.4(0.1)$ & $98.0(0.1)$\\
8 & $49.9$ & $79.1$ & $88.5\mid100$ & $82.7$ & $94.2(0.6)$ &  $91.7(1.1)$ & $94.9(0.6)$ & $94.0(0.2)$\\
9 & $34.0$ & $93.2$ & $96.5\mid100$ & $95.9$ & $98.1(0.2)$ &  $98.3(0.1)$ & $98.2(0.1)$ & $97.5(0.1)$\\
Avg & $51.7$ & $84.7$ & $91.6\mid100$ & $89.0$ &$93.7$  & $92.5$  & $92.6$ & $93.2$\\
\hline
\multicolumn{3}{l}{CIFAR-10}\\
0 & $41.1$ & $61.7(4.1)$ & $69.0\mid80.0$ & $81.7(0.2)$ & $78.0(0.3)$ & $63.8(2.3)$ &  $86.5(0.4)$ & $90.4(0.2)$\\
1 & $57.1$ & $65.9(2.1)$ & $44.2\mid90.5$ & $76.7(1.0)$ & $75.0(0.2)$ & $63.6(0.8)$ &  $95.3(0.1)$ & $95.9(0.1)$\\
2 & $53.8$ & $50.8(0.8)$ & $68.3\mid64.7$ & $66.7(1.0)$ & $68.1(0.5)$ & $60.0(1.7)$ & $75.1(0.3)$ & $81.2(0.4)$\\
3 & $51.2$ & $59.1(1.4)$ & $51.3\mid71.5$ & $67.1(1.5)$ & $66.7(0.4)$ & $62.3(2.1)$ & $78.8(0.7)$ & $84.9(0.3)$\\
4 & $52.2$ & $60.9(1.1)$ & $76.7\mid83.8$ & $73.6(2.0)$ & $77.9(0.8)$ & $69.4(1.5)$ & $89.2(0.3)$ & $91.8(0.2)$\\
5 & $49.3$ & $65.7(2.5)$ & $50.0\mid70.0$ & $74.4(2.0)$ & $68.6(0.3)$ & $61.6(1.1)$ & $87.0(0.4)$ & $90.3(0.2)$\\
6 & $64.9$ & $67.7(2.6)$ & $72.4\mid83.0$ & $74.4(0.9)$ & $81.2(0.4)$ & $73.0(2.3)$ & $88.9(0.9)$ & $92.1(0.3)$\\
7 & $55.3$ & $67.3(0.9)$ & $51.3\mid76.7$ & $74.3(0.2)$ &$74.8(0.2)$ & $61.2(1.0)$ & $91.1(0.4)$ & $93.3(0.3)$\\
8 & $51.9$ & $75.9(1.2)$ & $69.0\mid82.8$ & $80.0(1.7)$ & $79.1(1.0)$ & $72.3(2.3)$ & $92.4(0.3)$ & $94.7(0.1)$\\
9 & $54.2$ & $73.1(1.2)$ & $43.3\mid87.5$ & $76.2(0.7)$ & $76.1(0.2)$ & $66.5(1.5)$  & $95.8(0.1)$ &  $96.2(0.1)$\\
Avg & &$64.8$  &$59.5\mid79.1$  & $74.2$ & $74.5$ & $65.4$  & $88.0$ &  $91.1$\\
\hline
\end{tabular}
\end{table}

\end{document}